\documentclass[lettersize,journal]{IEEEtran}
\usepackage{amsmath, amssymb, amsfonts}
\usepackage{algorithmic}
\usepackage{algorithm}
\usepackage{array}
\usepackage[caption=false,font=normalsize,labelfont=sf,textfont=sf]{subfig}
\usepackage{textcomp}
\usepackage{stfloats}
\usepackage{url}
\usepackage{verbatim}
\usepackage{graphicx}
\usepackage{cite}
\usepackage[table]{xcolor}
\usepackage{xcolor}
\usepackage{float}
\usepackage{enumitem}
\usepackage{multirow}
\usepackage{newfloat}
\usepackage[table,dvipsnames,xcdraw]{xcolor}
\usepackage{listings}
\usepackage{booktabs}
\definecolor{forestgreen}{RGB}{50, 150, 50} 
\definecolor{darkblue}{rgb}{0, 0, 0.55} 
\begin{document}

\title{STAR: Semantic-Temporal Adaptive Representation Learning for Few-Shot Action Recognition}

\author{Hongli Liu,~\IEEEmembership{Student Member,~IEEE,}
Yu Wang,~\IEEEmembership{Member,~IEEE,}
Shengjie Zhao,~\IEEEmembership{Senior Member,~IEEE,}
\thanks{Hongli Liu, Yu Wang, and Shengjie Zhao are with the School of Computer Science and Technology, Tongji University, Shanghai, China, 201804, and with the Engineering Research Center of Key Software Technologies for Smart City Perception and Planning, Ministry of Education. (e-mail: hongli01@tongji.edu.cn;
yuwangtj@yeah.net; shengjiezhao@tongji.edu.cn)}
\thanks{Corresponding aunthors: Yu Wang, and Shengjie Zhao}
}
 
\markboth{ IEEE TRANSACTIONS ON CIRCUITS AND SYSTEMS FOR VIDEO TECHNOLOGY,,~Vol.~14, No.~8, August~2021}%
{Shell \MakeLowercase{\textit{et al.}}: A Sample Article Using IEEEtran.cls for IEEE Journals}
\IEEEpubidadjcol
\maketitle

\begin{abstract}
Few-shot action recognition (FSAR) requires models to generalize to novel action categories from only a handful of annotated samples. 
Despite progress with vision–language models, existing approaches still suffer from semantic–temporal misalignment, where static textual prompts fail to capture decisive visual cues that appear sparsely across sequences, and from inadequate modeling of multi-scale temporal dynamics, as short-term discriminative cues and long-range dependencies are often either oversmoothed or fragmented.
To address these challenges, we propose Semantic Temporal Adaptive Representation Learning (STAR), a unified framework, consisting of a semantic-alignment component and a temporal-aware component, effectively bridging the semantic and temporal gaps and transferring the sequence modeling capability of Mamba into the FSAR. 
The semantic alignment module introduces a Temporal Semantic Attention (TSA) mechanism, which performs frame-level cross-modal alignment with textual cues, ensuring fine-grained semantic--temporal consistency. 
The temporal-aware module incorporates a Semantic Temporal Prototype Refiner (STPR) that integrates semantic-guided Mamba blocks with multi-frequency temporal sampling and bidirectional state-space refinement, yielding semantically aligned prototypes with enhanced discriminative fidelity and temporal consistency.
Furthermore, temporally dependent class descriptors derived from large language models (LLMs) provide long-range semantic guidance.  
Extensive experiments on five FSAR benchmarks demonstrate the consistent superiority of STAR over state-of-the-art methods. For instance, STAR achieves up to 8.1\% and 6.7\% gains on the SSv2-Full and SSv2-Small datasets under the 1-shot setting, and 7.3\% on HMDB51, validating its effectiveness under limited supervision. The code is available at \url{https://github.com/HongliLiu1/STAR-main}.

\end{abstract}

\begin{IEEEkeywords}
Few-shot learning, action recognition, meta-learning, multimodal representation learning.
\end{IEEEkeywords}
\begin{figure}[t] 
    \centering
    \includegraphics[width=0.47\textwidth]{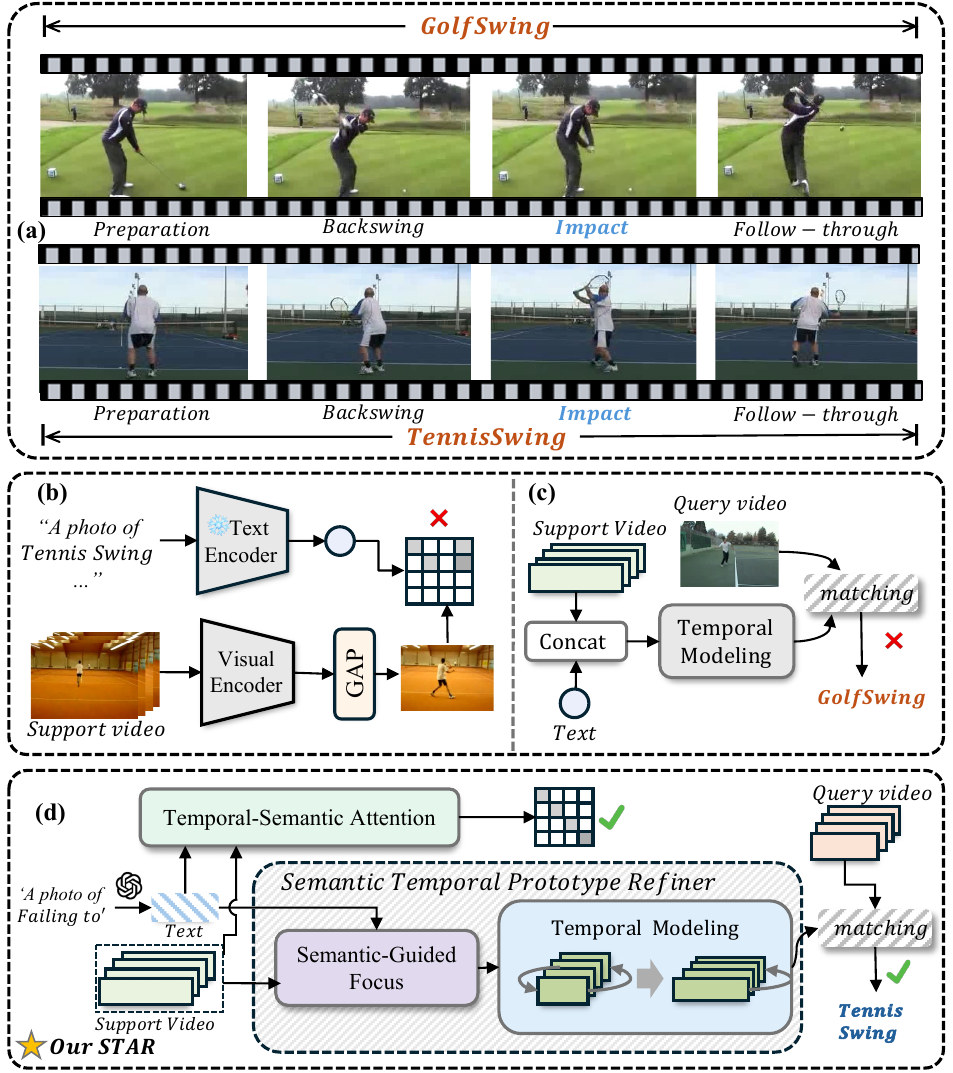}] 
    \caption{Motivation. (a) Actions such as \textit{Golf Swing} and  \textit{Tennis Swing}) differ only in a few decisive frames. (b) Average pooling compresses temporal structures, weakening semantic alignment. (c) Prior methods introduce semantic guidance but lack explicit semantic-temporal interactions. (d) Our STAR explicitly models frame-level semantic-temporal relations to precisely capture key action dynamics. }
    \label{fig:motivation} 
\end{figure}
\section{Introduction}
Video action recognition has emerged as a fundamental task in dynamic visual understanding, aiming to model complex temporal patterns and spatial interactions from sequential frames~\cite{9008827,tran2015learning,journals/corr/CarreiraZ17,9795869,arnab2021vivit}.
Despite the impressive progress of deep spatiotemporal networks, their reliance on large-scale frame-level annotations and high computational demand limits scalability in practical deployments.
Few-shot action recognition (FSAR) has therefore attracted increasing attention as a data-efficient alternative, enabling the recognition of unseen action categories from only a few annotated examples~\cite{zhu2018compound,cao2020few,wang2022hybrid}.
Most existing FSAR methods~\cite{snell2017prototypical,huang2024matching,bishay2019tarn,lee2025temporal} adopt a metric-based meta-learning framework, in which episodic training constructs class prototypes from support samples and classifies queries through similarity matching.
While effective under limited supervision, such approaches typically represent each video by a holistic clip-level embedding, overlooking the fine-grained temporal evolution and inter-instance relational cues that are crucial for action differentiation.
Consequently, models often depend excessively on background or contextual features, exhibit semantic bias, and fail to discriminate actions characterized by subtle motion variations.

The integration of contrastive vision–language models such as CLIP~\cite{radford2021learning} has introduced explicit semantic priors into few-shot action recognition (FSAR), alleviating the limitations of purely visual modeling. 
Recent CLIP-based extensions~\cite{wang2024clip} enhance generalization by aligning video embeddings with textual descriptions and narrowing the semantic–visual gap. Nevertheless, the semantic guidance offered by such models remains suboptimal due to inherent representational constraints. 
As shown in Fig.~\ref{fig:motivation}(a), actions such as \emph{Golf Swing} and \emph{Tennis Swing} exhibit highly similar appearances across most frames, whereas discriminative cues appear only in a brief impact phase, leading to severe inter-class confusion. 
Current methods~\cite{gao2021temporal} typically compress frame-level features through global average pooling to match the dimensionality of textual embeddings, as illustrated in Fig.~\ref{fig:motivation}(b). 
While this pooling strategy enables efficient cross-modal alignment, it simultaneously smooths out fine-grained motion variations critical for distinguishing temporally subtle actions, thereby weakening the correspondence between visual dynamics and semantic representations.
Following these limitations, subsequent studies have attempted to incorporate semantic priors into temporal representation learning through multi-scale matching or advanced temporal modules~\cite{cao2020few,huang2022compound,nie2024slowfocus}. 
However, approaches such as MVP-Shot~\cite{qu2024mvp} and TSAM~\cite{li2024frame} still couple visual and semantic features within a single temporal modeling block, lacking an explicit structure that establishes semantic–temporal correspondence, as depicted in Fig.~\ref{fig:motivation}(c). 
Moreover, most existing methods rely on static class-level descriptions, which fail to adapt to the evolving temporal phases of actions and thus struggle to capture transient dynamics.
These issues collectively weaken the effectiveness of semantic guidance during temporal alignment, especially in complex scenarios with semantic–visual conflicts.

To address these challenges, we propose Semantic–Temporal Adaptive Representation Learning (STAR), a unified framework that couples semantic alignment with hierarchical temporal refinement, and adapts the sequence modeling capabilities of Mamba~\cite{gu2023mamba} to FSAR under semantic guidance. As shown in Fig.~\ref{fig:motivation}(d), STAR comprises two complementary components: the Temporal Semantic Attention (TSA) module and the Semantic Temporal Prototype Refiner (STPR) module.  
In the TSA module, we introduce a frame-level cross attention mechanism to adaptively associate video features with key textual cues, facilitating fine-grained semantic-temporal alignment. Additionally, we construct temporally dependent class descriptors using large-scale pretrained language models, providing long-range semantic priors and enhancing the discriminative capacity of action representations.
Complementing the semantic alignment achieved by TSA, STPR progressively refines action prototypes through three coordinated mechanisms. The Semantic-Guided Focus (SGF) mechanism first enforces explicit correlation between textual cues and action-relevant segments, injecting semantic information into discriminative frames. 
Building on this, the Action-Specific Dynamic Temporal (ASD) submodule performs multi-frequency temporal sampling with unidirectional refinement, capturing localized motion dynamics and short-term variations under semantic guidance. 
The Action-Centric Unified Temporal (ACU) submodule employs bidirectional state-space modeling to integrate localized dynamics into a coherent global representation, maintaining long-range temporal dependencies and ensuring temporal stability. Through this hierarchical refinement, STPR produces prototypes that are both discriminative and temporally consistent.
Finally, we employ a frame-level temporal matching objective, compatible with standard aligners such as OTAM and Bi-MHM, to reduce semantic discrepancy between support and query videos and enable fine-grained action prediction.

The main contributions of this paper are summarized as follows:
\begin{enumerate}[label=\arabic*. , leftmargin=1.5em, itemsep=0.3em]
    \item We propose a novel STAR framework that performs top-down semantic-alignment and temporally-aware optimization to enhance the performance of FSAR.
    \item We design a Temporal Semantic Attention (TSA) module that performs frame-level video–text alignment via cross-attention and incorporates large language models (LLMs) to construct temporally dependent class descriptors, enhancing fine-grained semantic discrimination.
    \item To fully exploit the potential of semantic guidance, we introduce the Semantic Temporal Prototype Refiner (STPR) module, which explicitly models semantic-temporal associations and enables both hierarchical and global temporal modeling through semantic modulation. To the best of our knowledge, semantic priors have not yet been extensively explored in Mamba-based sequence modeling for FSAR.
    \item Extensive experiments on five FSAR benchmarks demonstrate that STAR establishes new state-of-the-art performance, highlighting its effectiveness and generalizability.
\end{enumerate}
\begin{figure*}[t] 
    \centering
    \includegraphics[width=0.95\textwidth]{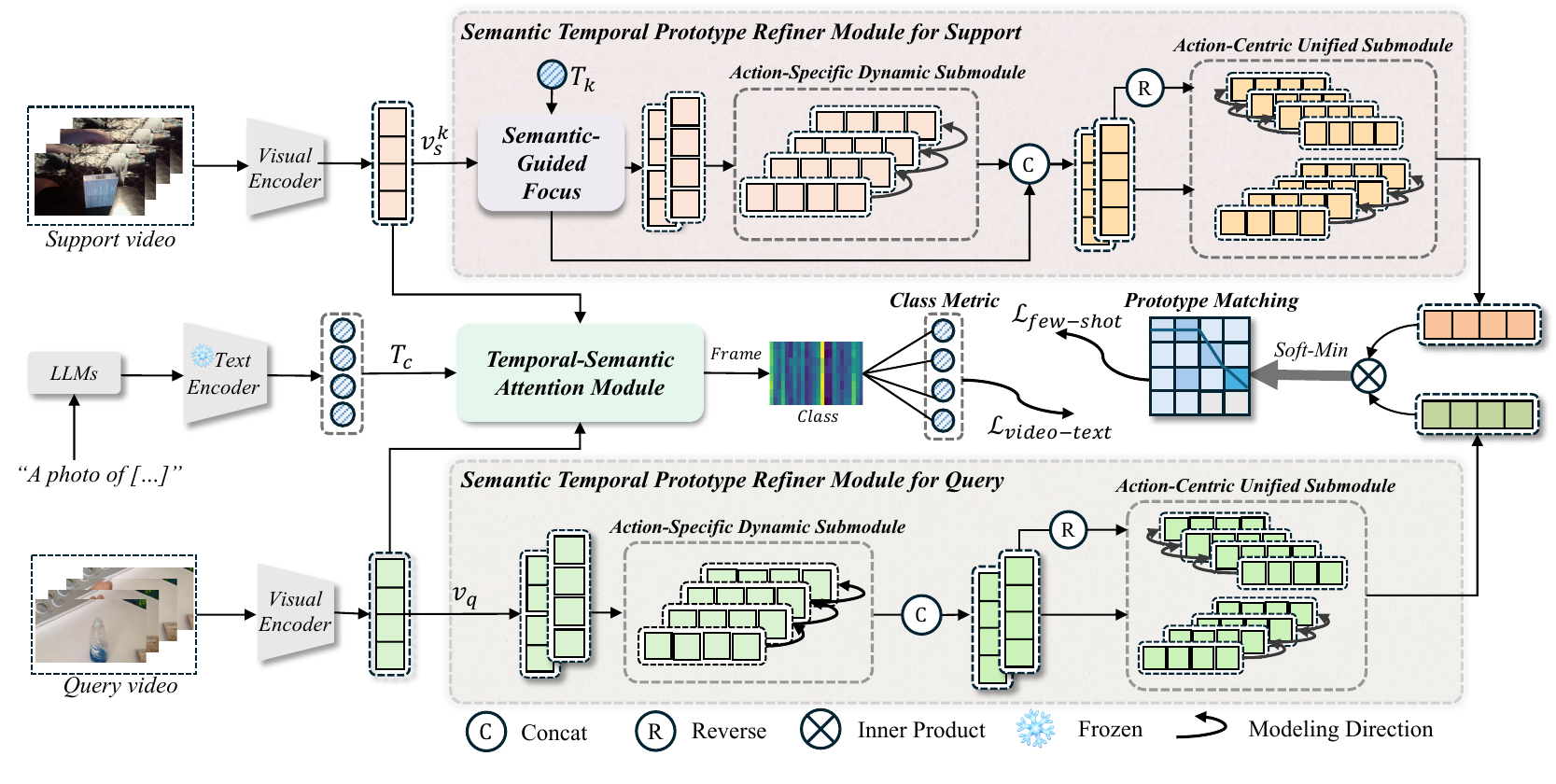}
    \caption{\textbf{Overall architecture of STAR.} Support and query videos are first encoded into frame-level features. The STPR module then applies semantically guided temporal modeling to capture long-range dynamics, followed by the TSA module which aligns the enriched frames with semantic descriptors. Finally, these matching signals are fused for query prediction. Additional support samples are omitted for clarity.}
    \label{fig:frame} 
\end{figure*}
\section{RELATED WORK}
\subsection{Few shot learning}
Few-shot learning (FSL) aims to recognize novel categories from only a few annotated samples, and has become a fundamental paradigm in computer vision~\cite{su2024reallocating,zhang2024few}.  
Existing approaches can be broadly categorized into metric-based and optimization-based frameworks.
Metric-based methods~\cite{vinyals2016matching,snell2017prototypical,sung2018learning,liu2026unify} learn a shared embedding space in which query and support samples are compared through distance metrics or learned similarity functions.
These methods are efficient and effective but often rely on the assumption that a single embedding space can generalize across tasks~\cite{jung2022few,lin2024revisiting}. 
To improve feature quality and adaptation, researchers have further explored attention-guided spatial features~\cite{tang2022learning} and meta-regularization strategies~\cite{tang2020blockmix}.
Optimization-based approaches, such as MAML~\cite{finn2017model}, instead meta-learn transferable initialization parameters that enable rapid adaptation to unseen tasks with limited updates.
While these paradigms have advanced image-level FSL, they largely assume static spatial representations and task-invariant embeddings~\cite{zhou2024meta,wang2024stability}. 
However, transferring such assumptions to video data neglects the inherently dynamic and compositional nature of human actions, where discriminative cues emerge in specific temporal segments rather than uniformly across frames.
This motivates the dedicated study of few-shot action recognition(FSAR), which goes beyond static spatial reasoning by requiring temporal modeling and, in recent works, cross-modal semantic integration.
\subsection{Few shot action recognition}
Few-shot action recognition (FSAR) extends FSL into the temporal domain, requiring models to recognize unseen action categories under severe data scarcity while capturing complex motion dynamics~\cite{wang2024few,cao2024task,lu2024cross,liu2023dual,liu2025motion}.
Early efforts~\cite{snell2017prototypical,vinyals2016matching,huang2024matching, bishay2019tarn,lee2025temporal} adopted temporal alignment-based metric learning, aligning query and support sequences at the frame level.
Representative methods such as CMN~\cite{zhu2018compound}, AmeFuNet~\cite{fu2020depth}, and OTAM~\cite{cao2020few} applied dynamic time warping to achieve fine-grained temporal correspondence. 
These approaches, however, focused solely on local alignment and lacked relational reasoning across instances. 
To capture richer spatiotemporal dependencies, later works introduced cross-video relational modeling and multi-scale temporal aggregation.
TRX~\cite{perrett2021temporal} incorporated cross-video attention to refine query features based on support context, and HyRSM~\cite{wang2022hybrid} proposed a hybrid relational strategy with bidirectional matching. SPRN~\cite{wang2021semantic} incorporates cross-frame spatial relational modeling to enhance spatial expressiveness. MoLo~\cite{wang2023molo} further explored multi-scale motion representation to handle varying temporal resolutions. 
Additionally, multi-view encoding and matching frameworks~\cite{tang2023m3net} have been proposed to effectively align fine-grained temporal patterns. 
Recent approaches are built upon vision-language pretrained models such as CLIP~\cite{radford2021learning}, which provide a strong cross-modal representation foundation~\cite{su2024reallocating,yin2024hierarchy,farina2025rethinking,tang2025connecting,tang2026cross}. 

MGCSM~\cite{yu2023multi} introduces a multi-speed global contextual subspace matching mechanism to generate robust, video-level action representations across varying temporal granularities. SA-CT~\cite{zhang2023importance} designs a spatial cross-attention module to align the spatial relations between query and support videos, mitigating the impact of spatial misalignment.With the incorporation of semantic priors, CLIP-FSAR~\cite{wang2024clip} constructs a video-text contrastive learning framework that integrates cross-modal knowledge from CLIP through a semantic prototype modulation mechanism, enhancing FSAR performance. 
Building on this, TSAM\cite{li2024frame} introduces sequence-aware adapters into the visual backbone to improve sensitivity to frame order. MVP-Shot~\cite{qu2024mvp} further proposes a multi-speed progressive alignment strategy to enhance generalization to varying motion rhythms.
Despite these advancements in temporal alignment and cross-modal representation learning, current methods still lack explicit semantic guidance for cross-modal temporal alignment. To address this gap, we propose a frame-level temporal-semantic alignment mechanism and a fine-grained semantic-guided temporal focus module to enable prototype learning that concentrates on the most discriminative action segments, thereby improving the alignment accuracy and generalization in few shot action recognition.

\subsection{State Space Model}
State Space Models (SSMs)\cite{gu2021combining,gu2021efficiently,wang2023selective} provide a principled framework for sequence modeling based on linear dynamical systems. 
By parameterizing continuous-time state transitions, SSMs offer strong inductive bias for long-horizon temporal continuity while maintaining linear computational complexity. The Structured State Space Sequence Model (S4)\cite{gu2021efficiently} demonstrates this capability by leveraging time-invariant system dynamics to effectively capture long-range dependencies.
Building upon this foundation, Mamba\cite{gu2023mamba,dao2024transformers} introduces a selective mechanism that adaptively filters input signals during state updates, enabling data-dependent information flow and enhancing modeling flexibility. This selective SSM architecture has achieved competitive or superior performance to Transformer-based models\cite{vaswani2017attention} across diverse benchmarks~\cite{yu2024mambaout}, offering an efficient alternative for long-context modeling. 
The visual variants of Mamba extend this paradigm to spatial and spatiotemporal domains.
Vision Mamba\cite{zhu2024visionmambaefficientvisual} incorporates bidirectional scanning to strengthen global spatial dependencies, while VMamba\cite{liu2024vmamba} generalizes this to four-directional context aggregation for richer visual representations.
For temporal modeling needs, VideoMamba~\cite{li2024videomamba} extends SSM to the domain of video analysis, demonstrating its effectiveness in extracting spatiotemporal features. 
Despite these advances, current Mamba-based architectures are primarily task-agnostic extensions, serving as generic sequence encoders without domain-specific structural adaptation~\cite{xie2025mamba,hatamizadeh2025mambavision,shaker2025groupmamba}.
In particular, the continuous global modeling scheme of Mamba lacks semantic constraints and localized selectivity, resulting in limited sensitivity to fine-grained motion cues. 
This deficiency becomes critical in FSAR, where discriminative semantics are often confined to sparse temporal segments and static global modeling may obscure decisive action phases. Addressing this limitation requires a semantically guided temporal refinement mechanism that preserves the SSM’s efficiency and long-range modeling capacity while adaptively emphasizing discriminative temporal regions.

\section{METHODS}
\subsection{Preliminaries}
In a standard few-shot learning protocol~\cite{cao2020few,perrett2021temporal}, the dataset is partitioned into disjoint training and test subsets, denoted as $D_{\text{train}}$ and $D_{\text{test}}$, respectively, with $D_{\text{train}} \cap D_{\text{test}} = \varnothing$. This partition guarantees evaluation on entirely unseen classes, thereby providing a rigorous measure of the model’s generalization capability.
FSAR adopts the episodic meta-learning paradigm~\cite{snell2017prototypical}, where the model is trained over a distribution of tasks rather than individual samples.
Episodic Learning~\cite{snell2017prototypical} to train the meta-learning model. During training, each episode task (i.e., a standard $N$-way $K$-shot task) involves randomly sampling $N$ classes from $D_{\text{train}}$, with the data from these classes divided into two sets:

\begin{itemize}[leftmargin=*, label=\textbullet,labelsep=0.5em]
  \item Support Set: $S = \{ (s_k, y_s^k) \}_{k=1}^{N \times K}$, where $y_s^k \in C_{\text{train}}$. Each class contributes $K$ labeled support videos providing limited supervision. $C_{\text{train}}$ denotes the class set in $D_{\text{train}}$.

  \item Query Set: $Q = \{ (q_p, y_q^p) \}_{p=1}^{P}$, where $y_q^p \in C_{\text{train}}$. Query samples are drawn from the remaining data of the same classes and are used to evaluate the model during each episode.
\end{itemize}
To prevent information leakage, the support and query sets are constructed to be disjoint, i.e., $S \cap Q = \emptyset$. The objective is to enable the model to learn from limited supervision in the support set and generalize to correctly classify the queries.
During inference, episodic tasks are constructed from $D_{\text{test}}$ using the same $N$-way $K$-shot protocol, and the trained model is evaluated on its ability to generalize to novel classes.
\subsection{Overall Architecture}

\begin{figure}[t] 
    \centering
    \includegraphics[width=0.46\textwidth]{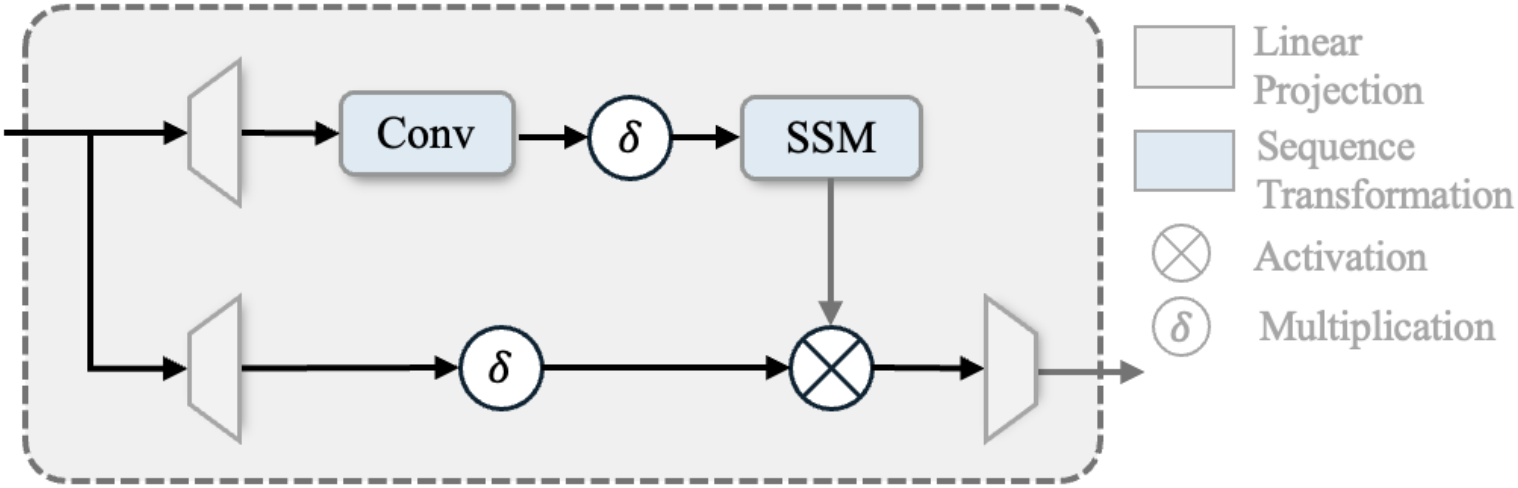}
    \caption{Structure of the Temporal State Space Module (TSSM), built upon the Mamba framework~\cite{gu2023mamba}, combining convolutional gating and state-space modeling for temporal sequence processing.} 
    \label{fig:ssm_m} 
\end{figure} 
The proposed STAR framework is illustrated in Fig.~\ref{fig:frame}. For clarity, we adopt an $N$-way 1-shot setting with only one query sample in the query set $Q$. STAR comprises two complementary modules designed to handle semantic–temporal challenges in FSAR. 
For each task, the support set $S$ and query sample $q$ are projected into a shared embedding space via the CLIP visual encoder $f_{\theta}(\cdot)$. The visual embeddings are given by:
\begin{equation}
V_s = \{ v_s^k \mid v_s^k = f_{\theta}(s_k) \}, \quad v_q = f_{\theta}(q),
\label{eq:feature_embedding}
\end{equation}
\noindent
where $k \in \{1, 2, \dots, N\}$, and $v_s^{k},v_q\in\mathbb{R}^{F\times D}$ denote per-frame embeddings with $F$ frames and feature dimension $D$.
Additionally, we leverage offline large language models (LLMs) to generate temporal-aware semantic descriptions for each training category, denoted as $C'_{\text{train}}$. Unlike conventional class names that only convey static category semantics, these LLM-generated descriptions encode phase-dependent motion cues and capture the temporal evolution of each action. 
Following CLIP’s original setting~\cite{radford2021learning}, we keep the text encoder $g_{\phi}(\cdot)$ frozen and use it to encode the semantic descriptions into category embeddings:
\begin{equation}
T_c = g_{\phi}(C'_{\text{train}}) = \{ c_1, c_2, \dots, c_M \},
\label{eq:textual_prototype}
\end{equation}
where $T_c \in \mathbb{R}^{M \times D}$ represents the set of category embeddings, with $M$ denoting the number of categories in $D_{\text{train}}$.

The Temporal-Semantic Attention (TSA) module aligns video frames with textual cues under the guidance of temporally dependent class descriptors, ensuring fine-grained semantic–temporal correspondence. 
Building upon this alignment, the Semantic Temporal Prototype Refiner (STPR) progressively synthesizes action prototypes that focus on discriminative temporal segments while preserving both localized motion dynamics and long-range temporal dependencies. 
By jointly optimizing asymmetric frame-level matching and cross-modal consistency, the framework produces representations that are semantically consistent, temporally coherent, and highly generalizable to unseen action categories.
\subsection{Temporal-Semantic Attention Module}
\begin{figure}[t]
    \centering
    \includegraphics[width=0.47\textwidth]{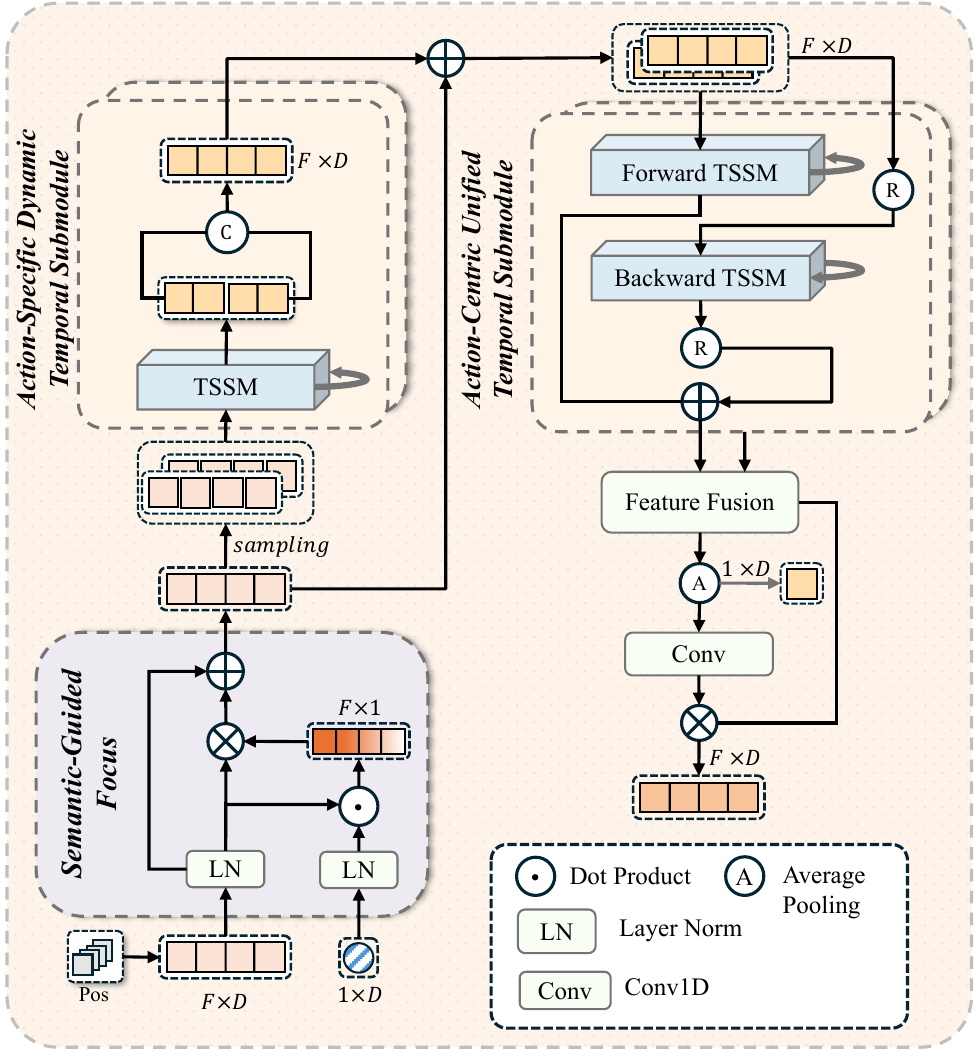}
    \caption{Illustration of the Semantic Temporal Prototype Refiner (STPR) module. STPR comprises the Semantic-Guided Focus (SGF), Action-Specific Dynamic Temporal (ASD), and Action-Centric Unified Temporal (ACU) submodules, which explicitly incorporate semantic information to facilitate both local and global temporal modeling.}
    \label{fig:subframe}
\end{figure}
We introduce the Temporal-Semantic Attention (TSA) module, which enhances semantic grounding under limited supervision through frame-wise cross-attention between class embeddings and video features, combined with contrastive alignment for semantic–temporal consistency. 
Given a support video’s action class embedding $T_k$ from the support set, we apply Layer Normalization to obtain $T_{\text{norm}}$, which serves as queries (Q), while the frame-level video features $V \in \{V_s, V_q\}$ act as keys (K) and values (V). Through multi-head cross attention, the semantic–temporal association weights are computed, yielding enhanced video representations:
\begin{align}
V_{\text{enhanced}} = \text{CrossAttention}(T_{\text{norm}}, V, V),
\end{align}
\noindent
where $V_{\text{enhanced}}^{(i)}\in \mathbb{R}^{M \times D}$ corresponds to the response of the video to the $i$-th class token, aggregating video information relevant to that class.

We adopt an InfoNCE-based contrastive objective that enforces cross-modal alignment between class-conditioned video responses and their textual descriptors:
{
\small
\begin{align}
\mathcal{L}_{\text{video-text}} = - \frac{1}{M} \sum_{i=1}^{M} \log \frac{\exp\big(\text{sim}(V_{\text{enhanced}}^{(i)},\ c_{(i)}) / \tau \big)}{\sum_{j=1}^{M} \exp\big(\text{sim}(V_{\text{enhanced}}^{(i)},\ c_{(j)}) / \tau \big)},
\end{align}
}
where $\text{sim}(\cdot,\cdot)$ denotes cosine similarity, $\tau$ is a learnable temperature, and $c_{(i)}$ corresponds to the ground-truth semantic embedding for the $i$-th class in the current episode.

Through cross-modal attention and contrastive alignment, TSA establishes fine-grained correspondence between video frames and textual cues, enforcing semantic–temporal consistency at the frame level. 
\subsection{Semantic Temporal Prototype Refiner Module}
To overcome the over-smoothing bias of state-space models and their weak alignment with textual cues under few-shot supervision, we design the Semantic Temporal Prototype Refiner (STPR).  
As illustrated in Fig.~\ref{fig:subframe}, the STPR addresses these issues through three submodules.  STPR refines class prototypes through a hierarchical process that couples semantic guidance with temporal modeling: the \emph{Semantic-Guided Focus (SGF)} highlights action-relevant frames via class-conditioned modulation; the \emph{Action-Specific Dynamic Temporal (ASD)} captures short-term motion variations through multi-frequency temporal refinement; and the \emph{Action-Centric Unified Temporal (ACU)} integrates these localized dynamics using bidirectional state-space modeling to maintain long-range coherence. This design yields temporally structured and semantically consistent prototypes that enhance generalization in few-shot action recognition.

\subsubsection{Temporal State Space Module (TSSM)} 
As shown in Fig.~\ref{fig:ssm_m}, we introduce the Temporal State Space Module (TSSM), a temporal modeling mechanism based on State Space Models (SSMs)~\cite{gu2021combining}.
The formulation of SSMs relies on linear ordinary differential equations (ODEs), expressed as:
\begin{equation}
h^{\prime}(t) = A h(t) + B x(t), \quad y(t) = C h(t),
\end{equation}
where $h(t) \in \mathbb{R}^{H}$ is the hidden state, $x(t) \in \mathbb{R}^{D}$ is the input, $y(t) \in \mathbb{R}^{D}$ is the output, and $A \in \mathbb{R}^{H \times H}, B \in \mathbb{R}^{H \times D}, C \in \mathbb{R}^{D \times H}$ are learnable parameters.

Following Mamba~\cite{gu2023mamba}, we discretize the continuous system for sequence inputs $x \in \mathbb{R}^{D \times L}$ with a learnable step size $\Delta$ (either scalar or per-step). Using the Zero-Order Hold (ZOH) method, the continuous matrices are converted to: 
\begin{equation}
\overline{A} = \exp(\Delta A), \quad 
\overline{B} = A^{-1}(\exp(\Delta A) - I)B.
\end{equation}

The hidden state and output are then updated as: 
\begin{equation}
h_t = \overline{A} h_{t-1} + \overline{B} x_t, \quad y_t = C h_t,
\end{equation}
where $h_t \in \mathbb{R}^{H}$ and $y_t \in \mathbb{R}^{D}$ for each time step $t$. 

Within this framework, TSSM employs the Selective Scan (S6)~\cite{gu2023mamba} mechanism, which parameterizes $\Delta$ and gating functions to achieve efficient sequence modeling while retaining the expressiveness of continuous SSMs.

\subsubsection{Semantic-Guided Focus (SGF) submodule}  
FSAR requires selectively emphasizing action-relevant frames while suppressing background or irrelevant segments, which cannot be achieved by vanilla temporal encoding alone.  
The Semantic-Guided Focus (SGF) addresses this issue by establishing class-conditioned, frame-level correspondence between text semantics and video features to temporally ground the evidence.

Specifically, the text embedding $T_k$ corresponding to the support video’s action class is normalized via Layer Normalization to obtain $T_{\text{norm}}$.
Dot-product similarity between $T_{\text{norm}}$ and each frame feature $v_s^k$ produces semantic relevance weights $W_{\text{video-text}} \in \mathbb{R}^{F}$, which highlight frames most consistent with class semantics.  
These weights reweight the support features through channel-wise broadcasting and element-wise scaling, and a residual connection is added to preserve original information:
\begin{equation}
\tilde{v}_s^k = v_s^k + \left( W_{\text{video-text}} \odot v_s^k \right),
\end{equation}
where $\odot$ denotes element-wise scaling with broadcast over the feature dimension.

This semantic reweighting provides a refined support representation that serves as a more reliable basis for subsequent temporal modeling and prototype refinement. In our default setting, SGF is applied only to the support video features $v_s^k$ during training to enhance semantic guidance. It is not used for the query video features $v_q$ in order to ensure fairness during evaluation.
\subsubsection{Action-Specific Dynamic Temporal (ASD) submodule}  
Following semantic–temporal alignment, the Action-Specific Dynamic Temporal (ASD) submodule models short-term motion at multiple temporal frequencies to suppress redundancy and capture fine-grained dynamics. 

Given a set of sampling strides $W$, for each stride $w \in W$, we define the sub-sequence sets of the support video $\tilde{v}_s^k$ and the query video $v_q$ as:
\begin{align}
V_{s}^{k,w} = \{v_w^{s_k, (o)}\}_{o=0}^{w-1}, \quad V_{q}^{w} = \{v_w^{q, (o)}\}_{o=0}^{w-1},
\label{eq:multi_stride_query}
\end{align}
where $o$ denotes the temporal offset. Each sub-sequence $v_w^{s_k, (o)}$ and $v_w^{q, (o)}$ lies in $\mathbb{R}^{T_w^{(o)} \times D}$, where $T_w^{(o)} = \left\lfloor \frac{F - o}{w} \right\rfloor$ represents the number of frames in the sub-sequence.

At a single temporal scale $w$, the ASD submodule independently processes the subsequence sets $V_{s}^{k,w}$ and $V_q^w$. Each subsequence is first passed through a unidirectional TSSM to generate local temporal representations
$\tilde{V}_{s}^{k,w}, \tilde{V}_{q}^{w} $. 
This step ensures causal modeling across each sub-sequence and enables accurate capture of fine-grained local temporal dynamics.

Since sub-sequences obtained with different offsets may have variable lengths, each is upsampled to $F$ frames by temporal interpolation. 
The resulting sequences are subsequently aggregated across offsets at corresponding time indices, yielding unified scale-specific features:
\begin{equation}
\begin{aligned}
\bar V_s^{k,w} &= \frac{1}{w}\sum_{o=0}^{w-1} \mathrm{Upsample}\!\big(\tilde v_w^{s_k,(o)}\big), \\
\hat{V}_q^w   &= \frac{1}{w}\sum_{o=0}^{w-1} \mathrm{Upsample}\!\big(\tilde v_w^{q,(o)}\big),
\end{aligned}
\label{eq:offset_avg}
\end{equation}
\noindent
where the aggregation is performed element-wise along the temporal dimension, yielding $\bar V_s^{k,w}, \bar V_q^{w}\in\mathbb{R}^{F\times D}$. 

For the support set, the aggregated feature $\bar V_s^{k,w}$ is further fused with the semantically enhanced representation $\tilde v_s^k$ obtained via SGF through residual addition:
\begin{equation}
\hat V_s^{k,w} = \bar V_s^{k,w} + \tilde v_s^k.
\end{equation}

This design allows the support features to benefit from both global semantic guidance and local temporal variations, while the query representation $\hat{V}_q^w$ remains unbiased by residual fusion, ensuring fair evaluation at inference.

\subsubsection{Action-Centric Unified Temporal (ACU) submodule}  
While ASD captures local temporal dynamics, it remains limited in modeling long-range dependencies, motivating the Action-Centric Unified Temporal (ACU) submodule for bidirectional global refinement. 

Given the scale-specific features $\hat{V}_s^{k,w}$ and $\hat{V}_q^w$, ACU applies a forward TSSM to the original sequences and a backward TSSM to their temporally reversed counterparts $\hat{V}_{s,rev}^{k,w}$ and $\hat{V}_{q,rev}^w$. 
The backward outputs are then reversed back to the original order and fused with the forward outputs, yielding representations that capture dependencies from both past and future contexts:
\begin{equation}
\begin{aligned}
\hat{V}_{s,\text{final}}^{k , w} &= \text{TSSM}_{\text{fwd}}(\hat{V}_s^{k , w}) + \text{rev}\left(\text{TSSM}_{\text{bwd}}(\hat{V}_{s,\text{rev}}^{k , w})\right),  \\
\hat{V}_{q,\text{final}}^{w} &= \text{TSSM}_{\text{fwd}}(\hat{V}_q^{w}) + \text{rev}\left(\text{TSSM}_{\text{bwd}}(\hat{V}_{q,\text{rev}}^{w})\right),
\end{aligned}
\label{eq:bi_tssm_fusion}
\end{equation}
where $\text{rev}(\cdot)$ represents the temporal reversal operation, ensuring that the backward output aligns with the forward output, enhancing the completeness and action relevance of the features.

To integrate the multi-frequency temporal dynamics captured by ASD, ACU further performs bidirectional refinement and aggregates representations across temporal scales into a unified feature sequence. Specifically, for each stride $w\!\in\!W$, ACU outputs refined features $\hat{V}_{s}^{k,w}$ and $\hat{V}_{q}^{w}$, which are fused through a weighted combination:
\begin{equation}
\hat{V}_{s,\text{fused}}^{k} = \sum_{w \in W} \alpha_w \hat{V}_{s,\text{final}}^{k , w}, \quad
\hat{V}_{q,\text{fused}} = \sum_{w \in W} \alpha_w \hat{V}_{q,\text{final}}^{w},
\label{eq:feature_fusion}
\end{equation}
where $\alpha_w = \tfrac{1}{|W|}$ denotes uniform weights across temporal resolutions. 

At last, we perform channel-wise recalibration based on global temporal context~\cite{liu2022spatial}. 
For the fused support and query features, we extract their global context descriptors, then apply a lightweight 1D convolution to model channel dependencies. The refined features are computed with a residual connection:
\begin{equation}
\begin{aligned}
\hat{V}_{s,\mathrm{refined}}^{k} &= \hat{V}_{s,\mathrm{fused}}^{k} + \hat{V}_{s,\mathrm{fused}}^{k} \odot \mathrm{Conv}^{1\mathrm{D}}\!\left(\hat{V}_{s,\mathrm{global}}^{k}\right), \\
\hat{V}_{q,\mathrm{refined}} &= \hat{V}_{q,\mathrm{fused}} + \hat{V}_{q,\mathrm{fused}} \odot \mathrm{Conv}^{1\mathrm{D}}\!\left(\hat{V}_{q,\mathrm{global}}\right),
\end{aligned}
\label{eq:eca}
\end{equation}
where $\hat{V}_{s,\mathrm{global}}^k$ and $\hat{V}_{q,\mathrm{global}}$ are the global descriptors obtained via the Temporal Global Average Pooling operation, $\mathrm{Conv}^{1\mathrm{D}}$ denotes 1D convolution, and $\odot$ is channel-wise multiplication. These resulting features preserve long-range temporal coherence while highlighting action-relevant semantics, enhancing prototype discriminability.

\subsection{Prototype Matching}
\begin{table*}[htbp]
\caption{Comparison of Recent Few-Shot Action Recognition Methods on SSv2-Small, SSv2-Full, HMDB51, UCF101, and Kinetics Under the 5-Way $K$-Shot Setting ($K=1$–$5$). Results Are Reported Using ResNet-50 and ViT-B Backbones Pretrained on ImageNet, Denoted as CLIP-RN50 and CLIP-ViT-B, Respectively. ``INet-RN50'' Refers to ResNet-50 Pretrained on ImageNet. Bold Numbers Indicate the Highest Accuracy, and ``–'' Denotes Results Not Reported in Existing Works.}
\setlength{\tabcolsep}{2pt}
\renewcommand{\arraystretch}{1.4}
\centering
\resizebox{\linewidth}{!}{
\begin{tabular}{c || c | c c |c c |c c |c c |c c }
\hline
\rowcolor{gray!20}
 & & \multicolumn{2}{c}{\textbf{SSv2-Small}} & \multicolumn{2}{c}{\textbf{SSv2-Full}} & \multicolumn{2}{c}{\textbf{HMDB51}} & \multicolumn{2}{c}{\textbf{UCF101}} & \multicolumn{2}{c}{\textbf{Kinetics}} \\
\rowcolor{gray!20}
\multirow{-2}{*}{\textbf{Method}} & \multirow{-2}{*}{\textbf{Pre-training}} & 1-shot & 5-shot & 1-shot & 5-shot & 1-shot & 5-shot & 1-shot & 5-shot & 1-shot & 5-shot \\
\hline \hline 
CMN~\cite{zhu2018compound} & INet-RN50 & 34.4 & 43.8 & 36.2 & 48.9 & - & - & - & - & 60.5 & 78.9 \\
OTAM~\cite{cao2020few} & INet-RN50 & - & - & 42.8 & 52.3 &  - &  - &  - & - & 73.0 & 85.8 \\
AmeFuNet~\cite{fu2020depth}& INet-RN50 & - & - & - & - & 60.2 & 75.5 & 85.1 & 95.5 & 74.1 & 86.8 \\
TRX~\cite{perrett2021temporal} & INet-RN50 & - & 59.1 & - & 64.6 & - & 75.6 & -&96.1 & 63.6 & 85.9 \\
SPRN~\cite{wang2021semantic}& INet-RN50 & - & - & - & - & 61.6 & 76.2 & 86.5 & 95.8 & 75.2 & 87.1 \\
HyRSM~\cite{wang2022hybrid} & INet-RN50 & 40.6 & 56.1 & 54.3 & 69.0 & 60.3 & 76.0 & 83.9 & 94.7 & 73.7 & 86.1  \\
MoLo~\cite{wang2023molo}& INet-RN50 & 41.9 & 56.2 & 55.0 & 69.6 & 60.8 &77.4 & 86.0 & 95.5 &74.0 & 85.6\\
MGCSM~\cite{yu2023multi} & INet-RN50 & - & - & - & - & 61.3 & 79.3 & 86.5 & 97.1 & 74.2 & 88.2 \\
SA-CT~\cite{zhang2023importance} & INet-RN50 & - & - & 48.9 & 69.1 & 60.4 & 78.3 & 85.4 & 96.4 & 71.9 & 87.1 \\
CLIP-FSAR~\cite{wang2024clip}& CLIP-RN50 & 52.1 & 55.8 & 58.7 & 62.8 & 69.4 & 80.7 & 92.4 & 97.0 & 90.1 & 92.0  \\
TSAM~\cite{li2024frame} & CLIP-RN50 & 53.1 & 58.0 & 60.2 & 64.3 & 72.9 & 82.8 & 93.5 & 97.5 & 93.0 & 93.5 \\
MVP-Shot~\cite{qu2024mvp} & CLIP-RN50 & 51.2 & 57.0 & - & - & 72.5 & 82.5 & 92.2 
 & 97.6 & 90.0 & 93.2\\
\textbf{STAR (Ours)} & CLIP-RN50 & \textbf{55.6}$_{\pm 0.4}$ & \textbf{59.2}$_{\pm 0.2}$ & \textbf{63.5}$_{\pm 0.3}$ & \textbf{67.1}$_{\pm 0.2}$ & \textbf{74.9}$_{\pm 0.3}$ & \textbf{85.0}$_{\pm 0.2}$ & \textbf{94.1}$_{\pm 0.2}$ & \textbf{98.2}$_{\pm 0.1}$ & \textbf{94.0}$_{\pm 0.3}$ & \textbf{95.6}$_{\pm 0.2}$ \\
\hline \hline 
SA-CT(ViT)~\cite{zhang2023importance}& INet-ViT-B & - & - & - & 66.3 & - & 81.6 & - & 98.0 & - & 91.2 \\
CLIP-FSAR~\cite{wang2024clip} & CLIP-ViT-B & 54.6 & 61.8 & 62.1 & 72.1 & 77.1 & 87.7 & 97.0 & 99.1 & 94.8 &95.4 \\
TSAM ~\cite{li2024frame} & CLIP-ViT-B & 60.5 & 66.7 & 65.8 & 74.3 & 84.5 & 88.9 & 98.3 & 99.3 & 96.4 & 97.2  \\
\textbf{STAR (Ours)} & CLIP-ViT-B & \textbf{61.2}$_{\pm 0.3}$ & \textbf{66.9}$_{\pm 0.3}$ & \textbf{66.1}$_{\pm 0.2}$ & \textbf{75.8}$_{\pm 0.3}$ & \textbf{86.0}$_{\pm 0.3}$ & \textbf{89.2}$_{\pm 0.2}$ & \textbf{98.5}$_{\pm 0.1}$ & \textbf{99.3}$_{\pm 0.1}$ & \textbf{97.3}$_{\pm 0.2}$ & \textbf{98.0}$_{\pm 0.1}$ \\
\hline 
\end{tabular}
}
\label{tab:fsar_methods}
\end{table*}

After temporal and semantic refinement through STPR, STAR performs metric-based matching in the prototype space to infer query categories. 
For a given category $c$, with $K \geq 1$ support samples, the prototype representation $U_c$ is the average of the refined support features from each sample in the support set~\cite{snell2017prototypical}.
Given the refined query representation $\hat{V}_{q}^{\text{fused}}$, its distance to the class prototype $U_c$ is measured:
\begin{align}
\text{dist}(q, c) = \text{Dist}(\hat{V}_{q,\text{refined}}, U_c),
\end{align}
\noindent
where $\text{Dist}$ uses the temporal metric from the OTAM~\cite{cao2020few} in STAR by default.

The probability distribution of the query video over the $N$ support categories is:
\begin{align}
P_{\text{few-shot}}(y=i \mid q) = \frac{\exp(-\text{dist}(q, c_i))}{\sum_{j=1}^{N} \exp(-\text{dist}(q, c_j))},
\label{eq:matching_dist}
\end{align}
\noindent
where the cross-entropy loss $\mathcal{L}_{\text{few-shot}}$ is used for optimization, consistent with previous work~\cite{wang2023molo,wang2024clip,qu2024mvp}.
The overall training objective jointly optimizes cross-modal semantic alignment and few-shot classification:
\begin{align}
\mathcal{L} = \mathcal{L}_{\text{video-text}} + \alpha \mathcal{L}_{\text{few-shot}},
\end{align}
\noindent
where $\alpha$ balances the contribution of the two terms. This joint objective strengthens both semantic–temporal consistency and discriminative capability under limited supervision.

At inference, we follow prior work~\cite{snell2017prototypical,perrett2021temporal,wang2024clip} and predict the query label as 
\(\hat y = \arg\max_{i} P_{\text{few-shot}}(y=i \mid q)\), 
which is equivalent to selecting the nearest class prototype under the OTAM distance defined in Eq.~\ref{eq:matching_dist}.

\section{Experiments}
\subsection{Datasets}
We evaluate our framework on five few-shot action recognition benchmarks: HMDB51~\cite{kuehne2011hmdb}, UCF101~\cite{soomro2012ucf101}, Kinetics~\cite{carreira2017quo}, and two variants of Something-Something V2 (SSv2)~\cite{goyal2017something}, namely SSv2-Full and SSv2-Small. 
All datasets are split into disjoint meta-training, meta-validation, and meta-testing label sets (\(\mathcal{C}_\text{train}\), \(\mathcal{C}_\text{val}\), \(\mathcal{C}_\text{test}\)), following established FSAR protocols.
HMDB51 contains 51 categories and 6{,}766 videos, with a 31/10/10 (train/val/test) class split~\cite{zhang2020few}. Its small scale and limited diversity make it a challenging benchmark. 
UCF101 consists of 101 categories with 13{,}320 videos, partitioned into 70/10/21 classes~\cite{zhang2020few}. 
Despite being larger, it exhibits strong background correlations, making it suitable for assessing robustness to contextual bias. 
Kinetics includes 100 categories, divided into 64/12/24 class split~\cite{zhu2018compound}. Many actions can be recognized via scene context, requiring robust temporal modeling.  
SSv2-Full is a large-scale dataset with 174 fine-grained actions and over 220K videos, split into 64/12/24 classes~\cite{cao2020few}. SSv2-Small follows the same split with fewer samples per class, providing a lightweight but temporally demanding alternative. Both emphasize object–action interactions and phase-sensitive temporal ordering.
All experiments follow the standard 5-way 1-shot and 5-way 5-shot protocols~\cite{cao2020few,wang2022hybrid,wang2024clip}, ensuring consistency with prior work.
\subsection{Implementation Details}
We evaluate STAR with two CLIP backbones: CLIP-RN50 (ResNet-50)~\cite{he2016deep} and CLIP-ViT-B/16~\cite{dosovitskiy2020image}. 
The text encoder is frozen to retain CLIP’s pre-trained knowledge, while temporal-aware semantic descriptions are generated offline using GPT-4 (ChatGPT)~\cite{achiam2023gpt} to provide semantic prompts emphasizing long-term temporal dependencies. 
The method-specific sampling frequencies are set to $W=\{1,2,4\}$ with a balancing factor $\alpha=1.5$ for the joint loss.
Following prior work~\cite{cao2020few,wang2024clip}, we uniformly sampled $T=8$ frames per video, resized each frame to $256\times256$, and randomly cropped it to $224\times224$. 
Standard data augmentations, including random horizontal flipping and color jittering, were applied~\cite{wang2022hybrid,qu2024mvp}. 
Optimization is performed using Adam~\cite{kingma2014adam} with an initial learning rate of $1\times10^{-4}$, decayed by a multi-step scheduler. 
Training lasts for 10 epochs, with 40,000 sampled tasks for SSv2 and 10,000 tasks for the other datasets. 
During evaluation, we report the mean classification accuracy over 10{,}000 randomly sampled 5-way 1-shot and 5-shot tasks. 
All experiments are conducted on two NVIDIA RTX A6000 GPUs (48GB). 
\subsection{Comparison with state-of-the-art methods}
We compare STAR with recent state-of-the-art approaches across five FSAR benchmarks under standard 5-way 1-shot and 5-shot protocols. Top-1 accuracy results are reported in Table~\ref{tab:fsar_methods}. STAR establishes new state-of-the-art performance across all datasets, validating the efficacy of our semantic-temporal modeling paradigm for few-shot action recognition.

Notably, STAR demonstrates substantial superiority on the temporally demanding SSv2 benchmarks. On SSv2-Small, our ViT-B-based model achieves 61.2\% and 66.9\% in 1-shot and 5-shot settings respectively, yielding gains over strong multimodal competitors like TSAM~\cite{li2024frame} and MVP-Shot~\cite{qu2024mvp}. This superiority extends to the larger SSv2-Full dataset, where STAR delivers notable absolute improvements of 5.8\% and 5.1\% in the 1-shot setting, alongside 4.3\% and 3.7\% in the 5-shot setting, for the CLIP-RN50 and CLIP-ViT-B backbones, respectively. 
While recent multimodal methods such as CLIP-FSAR~\cite{wang2024clip} improve upon single-modality baselines like OTAM~\cite{cao2020few}, their reliance on global average pooling (GAP) for semantic alignment severely compromises temporal precision. 
By contrast, STAR integrates frame-level semantics, successfully capturing discriminative motion cues and temporal ordering that pooling-based methods inherently fail to preserve.

Such robust representation capability translates consistently to other standard benchmarks. On HMDB51, STAR surpasses CLIP-FSAR by significant margins of 5.5\% and 4.3\% for 1-shot and 5-shot classification, respectively, underscoring its advantage in distinguishing visually analogous actions in data-scarce regimes. Given that HMDB51 features high intra-class variance and complex temporal dynamics with limited scene-specific context, the observed gains strongly indicate that our method successfully aligns semantic concepts with explicit motion trajectories. Furthermore, on UCF101 and Kinetics, where prominent spatial biases typically saturate baseline performance, STAR consistently yields absolute improvements of 1\% to 3\% over previous methods. Ultimately, these results confirm that our refinement mechanisms extract genuine temporal discrimination rather than merely exploiting static appearance priors, thereby demonstrating remarkable generalization across diverse visual domains. 

\begin{table}[t]
\begin{centering}  
\caption{Ablation results of key components in our STAR. We evaluate the contributions of the core modules: Temporal Class Representations (TCR), Temporal--Semantic Alignment (TSA), and Semantic--Temporal Prototype Refiner (STPR).}
\renewcommand{\arraystretch}{1.3}
\resizebox{\linewidth}{!}{ 
\begin{tabular}{c c c |c c|c c}
\hline
\rowcolor{gray!20}
 & & & \multicolumn{2}{c|}{\textbf{SSv2-Full}} & \multicolumn{2}{c}{\textbf{Kinetics}} \\
\rowcolor{gray!20}
\multirow{-2}{*}{\textbf{TCR}} & \multirow{-2}{*}{\textbf{TSA}} & \multirow{-2}{*}{\textbf{STPR}} & \textbf{1-shot} & \textbf{5-shot}  & \textbf{1-shot} & \textbf{5-shot} \\ \hline \hline
\textbf{$\times$} & \textbf{$\times$} & \textbf{$\times$} & 54.8 & 60.3& 82.1&86.9 \\ 
\textbf{\checkmark} & \textbf{$\times$} & \textbf{$\times$} & 56.0 & 61.3 & 83.3&88.1 \\ 
\textbf{$\times$} & \textbf{\checkmark} & \textbf{$\times$} & 55.3 & 60.9 &82.7 & 87.8 \\ 
\textbf{$\times$} & \textbf{$\times$} & \textbf{\checkmark} & 59.7 & 63.8 &84.1 & 89.0 \\ \hline
\textbf{\checkmark} & \textbf{\checkmark} & \textbf{$\times$} & 60.6 & 64.3&91.8&93.6 \\ \
\textbf{$\times$} & \textbf{\checkmark} & \textbf{\checkmark} & 62.7 & 66.4& 93.2&94.9  \\ 
\textbf{\checkmark} & \textbf{$\times$} & \textbf{\checkmark} & 63.0 & 66.5& 93.4&95.2  \\ 
\textbf{\checkmark} & \textbf{\checkmark} & \textbf{\checkmark} &  \textbf{63.5} &  \textbf{67.1} &  \textbf{94.0} & \textbf{95.6} \\ \hline
\end{tabular}
}
\label{tab:ablationresults}
\end{centering}
\end{table}
\begin{table}[t]
\centering
\caption{Ablation results of STPR sub-modules on the SSv2-Full dataset under 1-shot and 5-shot settings, with the sampling frequency set to $W = \{1, 2, 4\}$. 
$\Delta$ denotes the improvement compared to the ACU-only baseline.}
\renewcommand{\arraystretch}{1.3}
\begin{tabular}{c c c | c c | c c}
\hline
\rowcolor{gray!20}
\multicolumn{3}{c|}{\textbf{FPS}} & \multicolumn{2}{c|}{\textbf{SSv2-Full}} & \multicolumn{2}{c}{\boldmath$\Delta$} \\ 
\rowcolor{gray!20}
\textbf{SGF} & \textbf{ASD} & \textbf{ACU} & \textbf{1-shot} & \textbf{5-shot} & \textbf{1-shot} & \textbf{5-shot} \\ 
\hline \hline 
$\times$ & $\times$ & \checkmark & 61.4 & 65.2 & –   & –   \\ 
$\times$ & \checkmark & $\times$ & 61.3 & 65.6 & -0.1 & +0.4 \\ 
$\times$ & \checkmark & \checkmark & 62.2 & 66.4 & +0.8 & +1.2 \\ 
\checkmark & \checkmark & \checkmark & \textbf{63.5} & \textbf{67.1} & \textbf{+2.1} & \textbf{+1.9} \\ 
\hline
\end{tabular}
\label{tab:fps_ablation}
\end{table} 
These advancements are driven by the synergy of our proposed modules. The Temporal-Semantic Attention (TSA) module grounds video features into class-level semantics through fine-grained alignment, circumventing the information loss caused by uniform frame weighting. Simultaneously, the Semantic Temporal Prototype Refiner (STPR) leverages multi-frequency integration to model both local dynamics and long-range dependencies. Together, they synthesize temporally coherent prototypes, effectively bridging the semantic gap between static vision-language pretraining and dynamic video understanding.
\subsection{Ablation study}
To demonstrate the effectiveness of the proposed STAR framework, we conduct ablation studies to assess the contribution of each component. Unless otherwise specified, all experiments use CLIP ResNet-50 as the default visual encoder.
\subsubsection{Ablation on STAR Components}

As detailed in Tables~\ref{tab:ablationresults} and \ref{tab:fps_ablation}, we evaluate the individual contributions and interactions of STAR's core modules: Temporal Class Representations (TCR), Temporal Semantic Alignment (TSA), and the Semantic Temporal Prototype Refiner (STPR) on the SSv2 Full and Kinetics. We also analyze STPR's internal mechanisms by isolating its three submodules: Action Specific Dynamic Temporal (ASD), Action Centric Unified Temporal (ACU), and Semantic Guided Focus (SGF).

TCR introduces category-level temporal priors derived from offline descriptors, providing a moderate yet consistent improvement over the baseline. It enhances prototype stability by injecting long-range temporal structure, achieving 56.0\% and 61.3\% accuracy on SSv2-Full and 83.3\% and 88.1\% on Kinetics-100 under the 1-shot and 5-shot settings, respectively. When combined with TSA, these gains further expand to 60.6\% and 64.3\% on SSv2-Full, confirming that global temporal priors and frame-level semantic cues are highly complementary. 
TSA establishes explicit cross-modal correspondence at the frame level, mitigating the semantic--temporal misalignment inherent to global pooling. Although its standalone improvement remains moderate, it plays a crucial role in guiding temporal refinement. Integrated with either TCR or STPR, TSA consistently enhances the alignment between visual and semantic spaces, resulting in significant accuracy gains across datasets.  
STPR provides the most substantial contribution by transforming video features into temporally discriminative and semantically calibrated prototypes. Through multi-frequency refinement and bidirectional state-space modeling, STPR captures both local motion details and long-range dependencies, producing stable temporal prototypes that generalize effectively. 
When combined with TCR and TSA, STAR achieves its best overall performance, yielding 63.5\% and 67.1\% on SSv2-Full, and 94.0\% and 95.6\% on Kinetics-100. This demonstrates the effectiveness of integrating temporal priors, semantic alignment, and hierarchical refinement within a unified representation framework. 

Within STPR, ASD enhances sensitivity to short-term motion dynamics through multi-rate causal refinement, while ACU reinforces temporal coherence via bidirectional state-space aggregation. 
Their combination balances local fidelity and global consistency, and further the inclusion of SGF responsible for semantic focusing directs refinement toward action-relevant regions, yielding the highest prototype quality. Collectively, TCR, TSA, and STPR form a coherent semantic--temporal learning pipeline in which each component contributes distinct yet complementary strengths, enabling robust generalization in few-shot action recognition.

\subsubsection{Ablation on Multi-Frequency Temporal Sampling}
\begin{table}[t]
\centering
\caption{Ablation results of different sampling frequency configurations in the hierarchical temporal modeling module of STAR.}
\label{tab:multi-scale}
\resizebox{0.96\linewidth}{!}{  
\renewcommand{\arraystretch}{1.3}
\begin{tabular}{c|c c|c c }
\hline
\rowcolor{gray!20}
 & \multicolumn{2}{c|}{\textbf{SSv2-Full}} & \multicolumn{2}{c}{\textbf{Kinetics}} \\
\rowcolor{gray!20}
\multirow{-2}{*}{\textbf{Multi-Frequency}} & \textbf{1-shot} & \textbf{5-shot} & \textbf{1-shot} & \textbf{5-shot} \\ 
\hline \hline
$$W = \{1\}$$ & 61.4 & 64.6 & 90.3 & 92.5 \\ \hline
$$W = \{2\}$$ & 61.3 & 64.2 & 89.7 & 92.8 \\ \hline
$$W = \{4\}$$ & 61.2 & 64.5 & 90.0 & 92.1 \\ \hline
$$W = \{1, 2\}$$ & 62.1 & 65.5 & 91.6 & 93.4 \\ \hline
$$W = \{1, 4\}$$ & 62.5 & 65.9 & 92.5 & 93.0 \\ \hline
$$W = \{2, 4\}$$ & 62.7 & 66.1 & 92.3 & 94.7 \\ \hline
$$W = \{1, 2, 4\}$$ & \textbf{63.5} & \textbf{67.1} & \textbf{94.0} & \textbf{95.6} \\ \hline
\end{tabular}
}
\end{table}
\begin{table}[t]
\centering
\caption{Performance comparison under different temporal alignment strategies on the Kinetics dataset. 
$\Delta$ denotes the improvement compared to the baseline under the same alignment.}
\label{tab:matching}
\renewcommand{\arraystretch}{1.3}
\begin{tabular}{c|c c|c c}
\hline
\rowcolor{gray!20}
 & \multicolumn{2}{c|}{\textbf{1-shot}} & \multicolumn{2}{c}{\textbf{5-shot}} \\
\rowcolor{gray!20}
\multirow{-2}{*}{\textbf{Method}} & Acc. & $\Delta$ & Acc. & $\Delta$ \\
\hline \hline 
Bi-MHM~\cite{wang2022hybrid} & 77.5 & –    & 89.0 & –    \\
CLIP-FSAR (Bi-MHM)           & 87.7 & +10.2 & 92.1 & +3.1 \\
MVP-Shot (Bi-MHM)            & 89.4 & +11.9 & 92.9 & +3.9 \\
\textbf{Ours (Bi-MHM)}       & \textbf{93.6} & \textbf{+16.1} & \textbf{95.8} & \textbf{+6.8} \\
\hline
OTAM~\cite{wang2024clip}     & 81.9 & –    & 91.8 & –    \\
CLIP-FSAR (OTAM)             & 87.6 & +5.7 & 91.9 & +0.1 \\
MVP-Shot (OTAM)              & 90.0 & +8.1 & 93.2 & +1.4 \\
\textbf{Ours (OTAM)}         & \textbf{94.0} & \textbf{+12.1} & \textbf{95.6} & \textbf{+3.8} \\
\hline
\end{tabular}
\end{table}
\begin{figure}[t]
    \centering
    \includegraphics[width=0.98\linewidth]{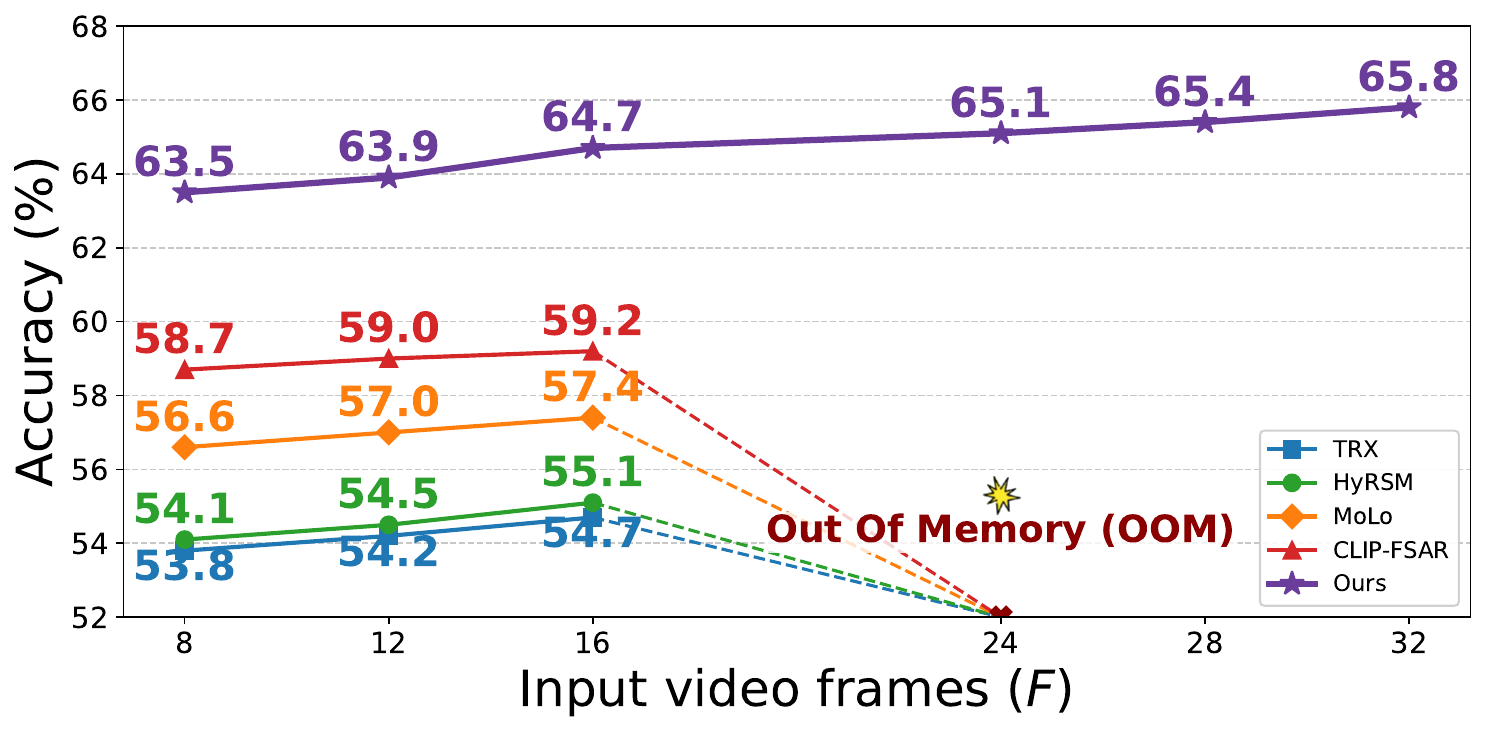}
    \caption{Top-1 accuracy (\%) comparison of different methods under varying frame lengths on SSv2-Full (1-shot).}
    \label{fig:frame_length_accuracy}
\end{figure}
\begin{figure}[t] 
    \centering
    \includegraphics[width=0.49\textwidth]{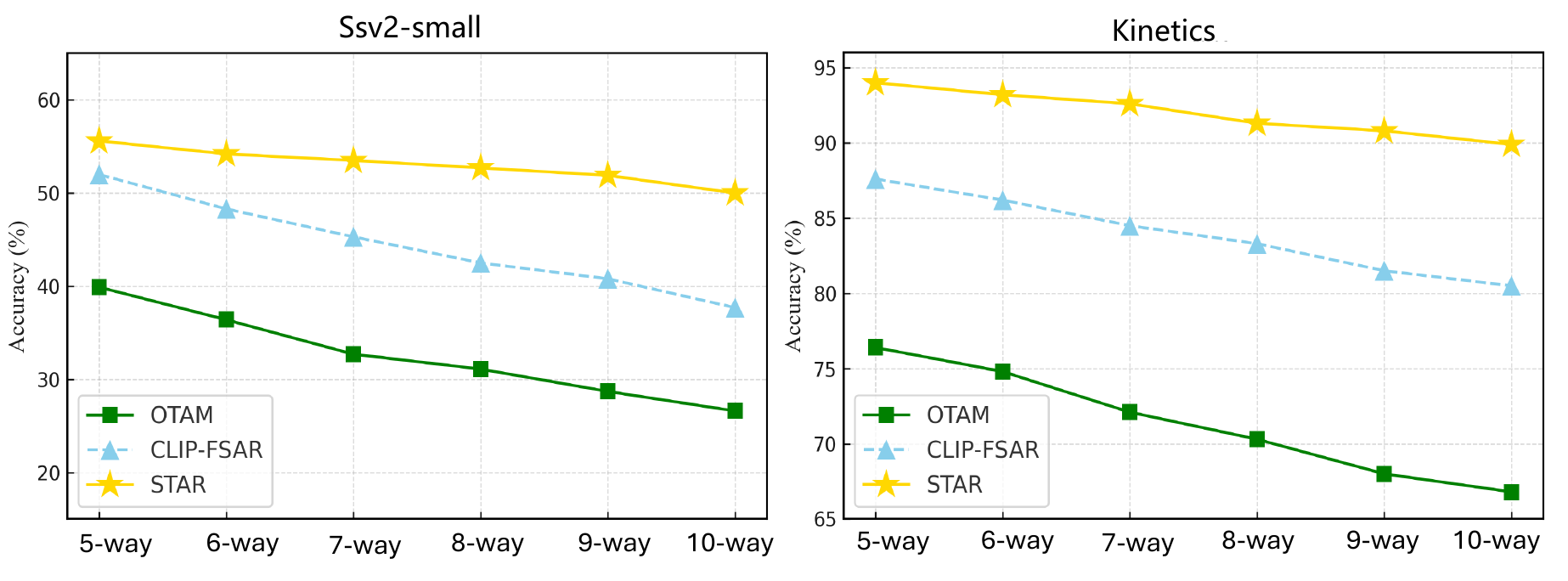} 
    \caption{N-way 1-shot performance of our STAR framework and baseline methods on SSv2-Small and Kinetics datasets, with $N$ varying from 5 to 10.} 
    \label{fig:nways}
\end{figure} 
We further investigate the effect of sampling frequency strategies within the hierarchical temporal modeling module of STAR. 
As summarized in Table~\ref{tab:multi-scale}, using a single sampling frequency ($W=\{1\}$, $\{2\}$, or $\{4\}$) leads to consistently lower accuracy on both datasets. 

This outcome suggests that a fixed temporal resolution restricts the model’s capacity to capture the full spectrum of motion dynamics: low-frequency sampling ($W=\{1\}$) overemphasizes coarse global context and tends to ignore subtle motion transitions, whereas high-frequency sampling ($W=\{4\}$) fragments long-term dependencies and exhibits higher sensitivity to temporal noise.  
In contrast, multi-frequency configurations substantially enhance performance. 
Dual-frequency setups such as $W=\{1,2\}$ and $W=\{2,4\}$ provide clear gains, indicating that complementary short- and long-range temporal cues jointly strengthen prototype representation. 
Notably, the $W=\{2,4\}$ setting achieves particularly strong results on Kinetics, reaching 92.3\% and 94.7\% under the 1-shot and 5-shot protocols, respectively, as high-frequency components are critical for capturing the rapid dynamics characteristic of large-scale actions.  
The complete configuration $W=\{1,2,4\}$ attains the best overall accuracy of 63.5\% and 67.1\% on SSv2-Full, and 94.0\% and 95.6\% on Kinetics. 
This confirms that integrating low-, medium-, and high-frequency temporal sampling yields a balanced and comprehensive representation: low frequencies capture holistic motion trends, medium frequencies preserve mid-range contextual information, and high frequencies resolve fine-grained variations. 
Such complementary coverage across temporal scales enhances temporal robustness, improves support–query alignment, and produces the most discriminative and stable prototypes.
\begin{figure*}[t]
    \centering
    \includegraphics[width=0.98\textwidth]{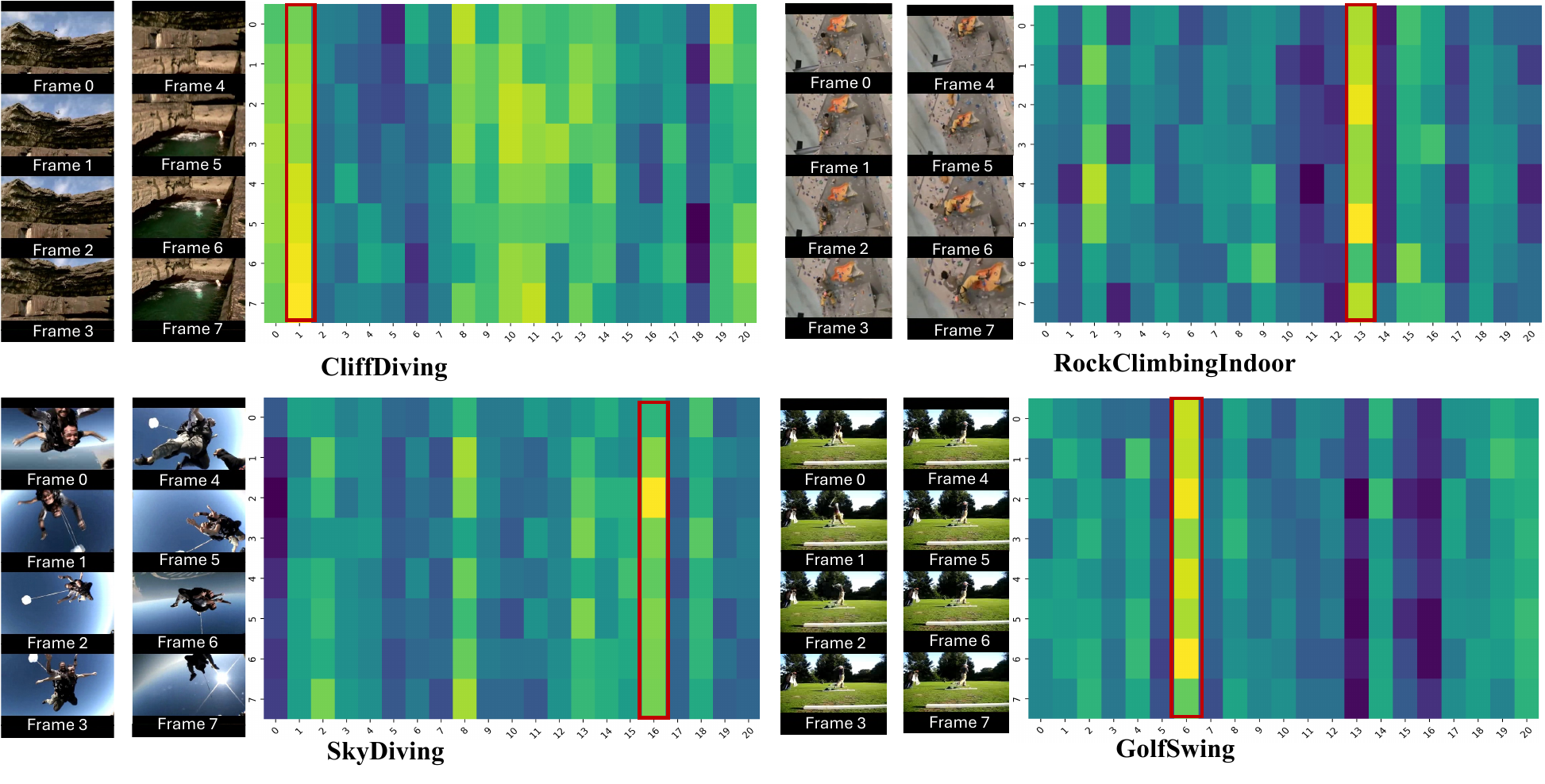} 
    \caption{Visualization of frame-to-class attention maps from the TSA module.The left side shows sampled video frames, while the right displays the corresponding heatmaps where brighter colors indicate stronger semantic alignment between frames (rows) and classes (columns).}
    \label{fig:tsa_attention}
\end{figure*}
\begin{table}[t]
\centering
\caption{Ablation of text prompt strategies on SSv2-Full and Kinetics. We compare manual templates with different LLM-generated semantic descriptors (TCR).}
\label{tab:llm_variants}
\small
\resizebox{1.0\linewidth}{!}{ 
\renewcommand{\arraystretch}{1.5}
\setlength{\tabcolsep}{8pt}
\begin{tabular}{c | c c | c c}
\hline
\rowcolor{gray!20}
 & \multicolumn{2}{c|}{\textbf{SSv2-Full (\%)}} & \multicolumn{2}{c}{\textbf{Kinetics (\%)}} \\
\rowcolor{gray!20}
\multirow{-2}{*}{\textbf{Prompt Strategy}} & \textbf{1-shot} & \textbf{5-shot} & \textbf{1-shot} & \textbf{5-shot} \\ 
\hline \hline
\texttt{"a photo of [CLS]"} & 62.7 & 66.4 & 93.2 & 94.9 \\ \hline
\texttt{Gemini-2.5-Pro} & \textbf{63.8} & \textbf{67.3} & 93.5 & 95.5 \\ \hline
\texttt{GPT-4o} & 63.5 & 67.1 & \textbf{94.0} & \textbf{95.6} \\ \hline
\end{tabular}
}
\end{table}
\begin{table}[t]
\centering
\caption{Complexity analysis for 5-way 1-shot HMDB51 and Kinetics evaluation. Here, we report Total  Params and Temporal Params(T-Params), FLOPS, FPS, and accuracy for each model.}
\label{tab:temporal_modules_analysis}
\small
\resizebox{1.0\linewidth}{!}{
\renewcommand{\arraystretch}{1.7}
\begin{tabular}{c | c c | c c | c c}
\hline
\rowcolor{gray!20}
\textbf{Method} & 
\textbf{\begin{tabular}[c]{@{}c@{}}Param \\ \end{tabular}} & 
\textbf{\begin{tabular}[c]{@{}c@{}}T-Params\end{tabular}} & 
\textbf{\begin{tabular}[c]{@{}c@{}}FLOPS \\\end{tabular}} & 
\textbf{\begin{tabular}[c]{@{}c@{}}FPS \\ \end{tabular}} &
\textbf{\begin{tabular}[c]{@{}c@{}}HMDB51 \\\end{tabular}} & 
\textbf{\begin{tabular}[c]{@{}c@{}}Kinetics \\ \end{tabular}} \\ 
\hline \hline
CLIP-FSAR & 71.90 & 25.10 & 468.3 & 3.35 & 69.4 & 90.1 \\ \hline
STAR (Ours) & 55.64 & 8.92 & 464.0 & 3.68 & 74.9 & 94.0 \\ \hline
\end{tabular}
}
\end{table}
\begin{figure}[t]
    \centering
\includegraphics[width=0.47\textwidth]{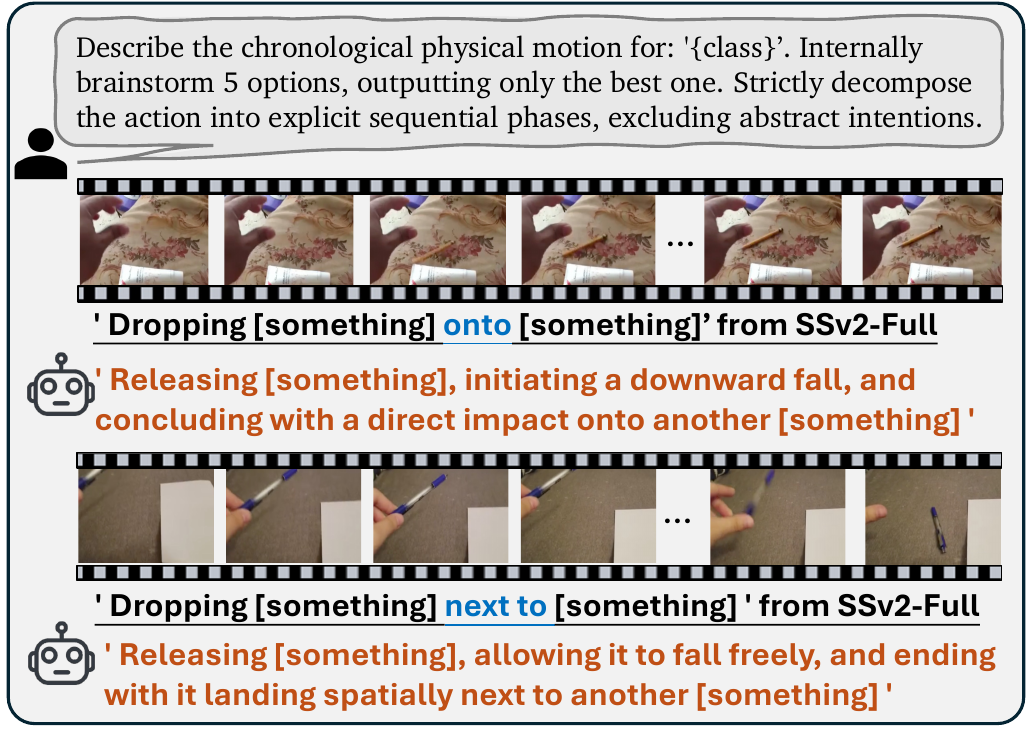} 
    \caption{Generation of Temporal Class Representations via LLM prompting. The diagram demonstrates how brief action class labels are transformed into explicit, sequential motion descriptions using a guided LLM prompt.}
    \label{fig:llm_prompt}
\end{figure}
\subsubsection{Influence of Different Alignment Strategies} 
To further examine the adaptability of STAR, we integrate it with two representative temporal alignment frameworks: Bi-MHM~\cite{wang2022hybrid}, which performs multi-hypothesis matching across subsequences, and OTAM~\cite{cao2020few}, which achieves optimal temporal alignment via dynamic programming. 
As reported in Table~\ref{tab:matching}, STAR consistently improves upon CLIP-FSAR and MVP-Shot across both settings. 
When combined with Bi-MHM, STAR attains 93.6\% and 95.8\% accuracy under the 1-shot and 5-shot protocols, surpassing CLIP-FSAR by 5.9 and 3.7 percentage points, respectively. 
With OTAM, it reaches 94.0\% and 95.6\%, outperforming CLIP-FSAR by 6.4 and 3.7 percentage points. 
These consistent gains demonstrate that STAR operates orthogonally to the underlying alignment mechanism: the integration of semantic guidance and multi-scale temporal refinement alleviates Bi-MHM’s dependence on local segment matching and complements OTAM’s global path optimization. 
Overall, STAR generalizes effectively across heterogeneous temporal alignment paradigms, confirming its robustness and flexibility as a plug-and-play enhancement for future few-shot action recognition frameworks.
\subsubsection{Temporal Scalability Analysis} 
\begin{figure*}[t] 
    \centering
    \includegraphics[width=\textwidth]{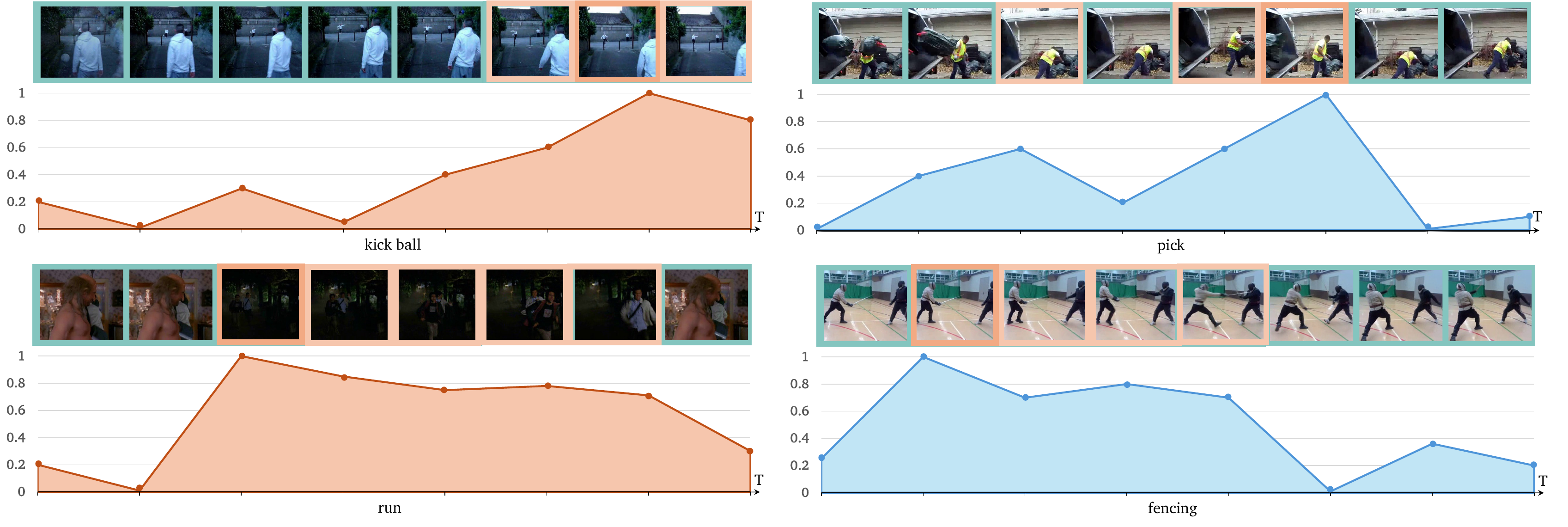} 
    \caption{Visualization of frame-level importance scores on the HMDB51 dataset. The line graphs illustrate the temporal distribution of importance scores across four action categories. The highlighted orange frames align with the score peaks, demonstrating our model's ability to localize key motion dynamics.}
    \label{fig:time_class}
\end{figure*}
We further evaluate the scalability of the proposed STAR model by extending the input sequence up to $F=32$ frames on the SSv2-Full dataset. As illustrated in Figure~\ref{fig:frame_length_accuracy}, traditional baseline methods such as TRX~\cite{perrett2021temporal}, HyRSM~\cite{wang2022hybrid}, MoLo~\cite{wang2023molo}, and CLIP-FSAR~\cite{wang2024clip} exhibit limited performance gains as the frame count increases. More importantly, these attention-based architectures typically encounter Out-Of-Memory (OOM) failures when reaching 24 frames, highlighting the prohibitive quadratic memory complexity inherent in standard self-attention. 
In contrast, our STAR model demonstrates more favorable linear scaling properties by effectively alleviating these computational constraints. This efficiency allows STAR to consistently leverage richer temporal information, with Top-1 accuracy rising steadily from 63.5\% at 8 frames to a peak of 65.8\% at 32 frames. These results suggest that the SSM-based architecture within the STPR module can efficiently capture long-range dependencies without the excessive overhead typical of Transformer-based models, providing a more scalable solution for high-fidelity video recognition.
\subsubsection{Performance of STAR in N-Way FSAR Tasks}
We further evaluate the generalization ability of STAR under more challenging N-way 1-shot settings on SSv2-Small and Kinetics, with $N$ varying from 5 to 10 (Fig.~\ref{fig:nways}). 
As $N$ increases, accuracy inevitably drops for all methods because the number of candidate classes grows while the support per class remains fixed, making discrimination harder and increasing the likelihood of class confusion. 
Nevertheless, STAR consistently achieves higher accuracy and exhibits a slower degradation compared with baseline methods. 
On SSv2-Small, where fine-grained action distinctions amplify the challenge, STAR shows a substantially smaller decline than CLIP-FSAR, confirming that its semantic–temporal refinement yields more robust prototypes under class-dense conditions. 
On Kinetics, although appearance cues reduce the relative difficulty, STAR still maintains the most stable performance, demonstrating its ability to suppress background bias and preserve discriminability as task complexity scales. 
These results highlight STAR’s scalability and robustness, ensuring reliable generalization even as the number of novel classes increases.
\begin{figure}[t] 
    \centering
    \includegraphics[width=0.47\textwidth]{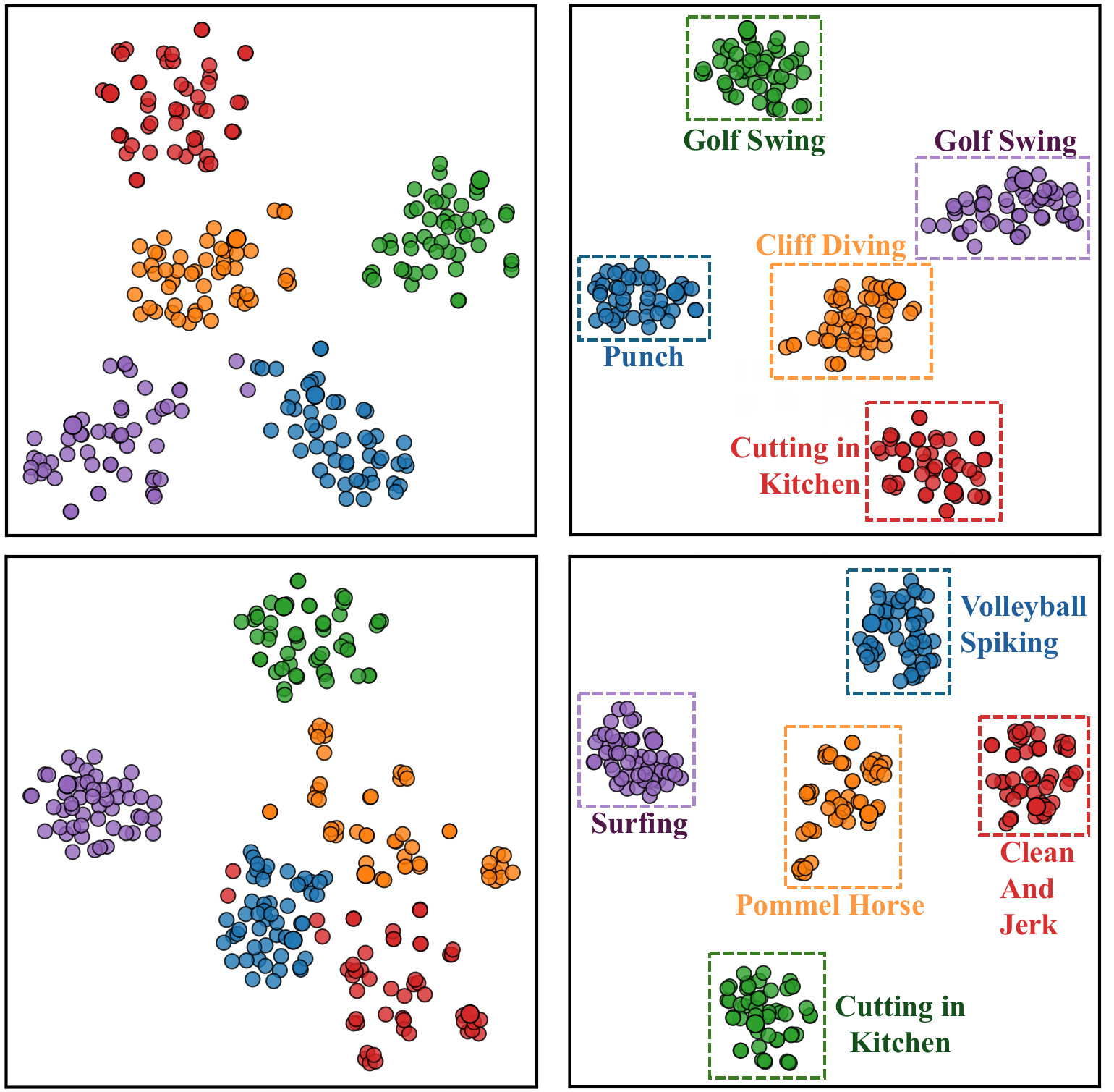}
    \caption{T-SNE visualization of five challenging action classes from the UCF101 test set under the 5-way 50-shot setting.} 
    \label{fig:tsne_matching} 
\end{figure} 
\subsubsection{Impact of LLM-Generated Descriptors}
We employ Large Language Models (LLMs) to generate Temporal Class Representations (TCR) as visualized in Fig.~\ref{fig:llm_prompt}, serving as prompts to guide temporal modeling. 
To evaluate the effectiveness and robustness of STAR to descriptor variations, we compare the traditional static template ``a photo of [CLS]'' against temporal prompts generated by two representative LLMs: GPT-4o~\cite{openai2024gpt4o} and Gemini-2.5-Pro~\cite{deepmind2025gemini}. These models differ inherently in their training data, generation styles, and linguistic abstraction.  As shown in Table~\ref{tab:llm_variants}, substituting the static template with LLM-generated temporally aware descriptions yields consistent and significant improvements across both datasets. For instance, on the temporally heavy SSv2-Full dataset, Gemini-2.5-Pro achieves the highest accuracy, outperforming the static baseline by 1.1\% in the 1-shot setting. Conversely, on the Kinetics dataset, GPT-4o delivers the optimal results, reaching 94.0\% and 95.6\% in the 1-shot and 5-shot settings, respectively. Despite minor dataset-specific preferences, STAR maintains stable performance across both LLM variants. This demonstrates that our semantic alignment mechanism effectively leverages rich temporal priors from advanced LLMs, remaining resilient to specific prompt phrasing and model-inherent biases.
\subsubsection{Model Efficiency Analysis} 
To analyze the effectiveness of training and inference, we provide a detailed comparison with the CLIP-FSAR in terms of parameters, FLOPs, and inference speed (FPS). For a fair comparison, we adopt the same visual encoder for both methods. As shown in Table~\ref{tab:temporal_modules_analysis}, compared to SOTA CLIP-FSAR, our model not only avoids additional overhead but actually reduces the overall computational burden, operating with fewer parameters and a faster inference speed. Despite this lightweight design, our STAR can bring 5.5\% and 3.9\% accuracy improvements on HMDB51 and Kinetics over CLIP-FSAR, respectively. These results demonstrate that our Mamba-based temporal architecture is both effective and efficient, offering promising performance improvements without sacrificing computational feasibility.
\subsection{Visualization}
\subsubsection{TSA Attention Map Visualization}  
To further interpret how TSA performs temporal--semantic alignment, we visualize the class--frame attention distributions on UCF101 under the 5-way 1-shot setting, as shown in Fig.~\ref{fig:tsa_attention}. 
Each panel displays sampled query frames (indices 0--7) on the left and the corresponding attention map on the right, where rows correspond to candidate action classes and columns to video frames. 
Color intensity encodes the degree of semantic alignment between each frame and its class embedding, thereby revealing how TSA evaluates the contribution of individual frames to class discrimination. 
The visualization reveals distinct alignment patterns across action types. 
For \textit{CliffDiving}, strong attention responses concentrate on the early frames depicting the jump and descent, while static background frames remain consistently suppressed. 
In \textit{RockClimbingIndoor}, TSA emphasizes frames capturing stable body postures against the climbing wall and down-weights transitional segments. 
For \textit{SkyDiving}, the attention for the ground-truth class peaks at mid-sequence frames showing the free-fall phase,  whereas sky-only frames exhibit weak activation. 
In \textit{GolfSwing}, attention remains uniformly high throughout the sequence but reaches maximum intensity during the downswing--impact interval, which contains the most discriminative motion cues. 
By anchoring frame-level representations to class-level semantics, TSA achieves precise temporal localization and ensures that prototype refinement is driven by action-relevant evidence, highlighting its essential role within the STAR framework.

\subsubsection{Visualization of Temporal Dynamics}  
We qualitatively assess the temporal grounding capability of our framework by visualizing the frame-level importance scores generated by STAR on the HMDB51 dataset (Fig.~\ref{fig:time_class}). Unlike standard temporal pooling that treats all frames uniformly, our model dynamically modulates temporal weights, effectively isolating core action dynamics from uninformative background contexts. For instance, in the \textit{run} sequence, which contains misleading narrative cutaways, STAR assigns near-zero weights to these irrelevant scene transitions, strictly attending to the active running motion. Furthermore, the generated attention trajectories align precisely with the kinetic evolution of the specific actions. In discrete actions such as \textit{kick ball} and \textit{pick}, the importance weights remain suppressed during the preliminary approach phases, but peak sharply at the exact moment of physical impact or active lifting. Conversely, for continuous interactions like \textit{fencing}, an elevated attention span is consistently sustained throughout the entire engagement phase. By explicitly filtering temporal noise and background interference, STAR distills highly purified video representations. This dynamic filtering mechanism significantly mitigates prototype pollution, ensuring that the subsequent few-shot semantic alignment is primarily driven by highly discriminative temporal cues.

\subsubsection{t-SNE Visualization} 
To qualitatively evaluate the discriminative capability of the learned representations, we perform t-SNE visualization on UCF101 (5-way 50-shot) and compare our STAR framework with the strong few-shot baseline CLIP-FSAR. As shown in Fig.~\ref{fig:tsne_matching}, although CLIP-FSAR (left panels) forms coarse category-level clusters, its embedding space remains poorly structured, exhibiting pronounced intra-class dispersion and blurred inter-class boundaries, particularly among semantically similar actions such as \textit{Golf Swing} versus \textit{Punch} and \textit{Volleyball Spiking} versus \textit{Clean and Jerk}, where scattered distributions and outliers frequently intrude into neighboring class regions, indicating that conventional attention-based adaptation is insufficient to suppress non-discriminative temporal noise. In contrast, STAR (right panels) produces a markedly more organized latent topology, characterized by compact intra-class distributions and clear inter-class separability, with challenging categories such as \textit{Cutting in Kitchen} and \textit{Pommel Horse} forming dense and well-isolated manifolds. 
This transition from dispersed to compact representations highlights that STAR enforces a more coherent and discriminative embedding space, enabling robust few-shot generalization under complex temporal dynamics.
\section{Conclusion}
We presented STAR, a semantic–temporal adaptive representation framework for few-shot action recognition. 
STAR couples semantic prompt guidance with multi-scale temporal modeling to address limited supervision and complex motion dynamics. 
The proposed TSA module enforces fine-grained semantic–temporal alignment, while the STPR module refines prototypes through multi-frequency temporal reasoning and semantic focusing. 
Working in tandem, these components enable precise temporal localization and discriminative prototype learning. 
Extensive experiments across multiple benchmarks verify STAR’s superiority over existing approaches, underscoring the importance of semantic–temporal synergy for generalizable video understanding.

\section{Acknowledgements}
This work was supported in part by the National Key Research and Development Project under Grant 2023YFC3806000, in part by the National Natural Science Foundation of China under Grant 62406226,  in part sponsored by Shanghai Sailing Program under Grant 24YF2748700, in part by New-Generation Information Technology under the Shanghai Key Technology R\&D Program under Grant 25511103500.
\bibliographystyle{IEEEtran}
\bibliography{ref}

@article{journals/corr/CarreiraZ17,
  author       = {Jo{\~{a}}o Carreira and
                  Andrew Zisserman},
  title        = {Quo Vadis, Action Recognition? {A} New Model and the Kinetics Dataset},
  journal      = {CoRR},
  volume       = {abs/1705.07750},
  year         = {2017},
  url          = {http://arxiv.org/abs/1705.07750},
  eprinttype    = {arXiv},
  eprint       = {1705.07750},
  bibsource    = {dblp computer science bibliography, https://dblp.org}
}

@inproceedings{sung2018learning,
  title={Learning to compare: Relation network for few-shot learning},
  author={Sung, Flood and Yang, Yongxin and Zhang, Li and Xiang, Tao and Torr, Philip HS and Hospedales, Timothy M},
  booktitle={IEEE conference on computer vision and pattern recognition},
  pages={1199--1208},
  year={2018}
}

@inproceedings{finn2017model,
  title={Model-agnostic meta-learning for fast adaptation of deep networks},
  author={Finn, Chelsea and Abbeel, Pieter and Levine, Sergey},
  booktitle={International conference on machine learning},
  pages={1126--1135},
  year={2017},
  organization={PMLR}
}

@inproceedings{arnab2021vivit,
  title={Vivit: A video vision transformer},
  author={Arnab, Anurag and Dehghani, Mostafa and Heigold, Georg and Sun, Chen and Lu{\v{c}}i{\'c}, Mario and Schmid, Cordelia},
  booktitle={IEEE/CVF International Conference on Computer Vision},
  pages={6836--6846},
  year={2021}
}

@inproceedings{zhu2018compound,
  title={Compound memory networks for few-shot video classification},
  author={Zhu, Linchao and Yang, Yi},
  booktitle={European conference on computer vision},
  pages={751--766},
  year={2018}
}

@INPROCEEDINGS{cao2020few,
  author={Cao, Kaidi and Ji, Jingwei and Cao, Zhangjie and Chang, Chien-Yi and Niebles, Juan Carlos},
  booktitle={2020 IEEE/CVF Conference on Computer Vision and Pattern Recognition}, 
  title={Few-Shot Video Classification via Temporal Alignment}, 
  year={2020},
  volume={},
  number={},
  pages={10615-10624},}

@INPROCEEDINGS{9008827,
  author={Lin, Ji and Gan, Chuang and Han, Song},
  booktitle={2019 IEEE/CVF International Conference on Computer Vision}, 
  title={TSM: Temporal Shift Module for Efficient Video Understanding}, 
  year={2019},
  volume={},
  number={},
  pages={7082-7092},
  }

@inproceedings{wang2022hybrid,
  title={Hybrid relation guided set matching for few-shot action recognition},
  author={Wang, Xiang and Zhang, Shiwei and Qing, Zhiwu and Tang, Mingqian and Zuo, Zhengrong and Gao, Changxin and Jin, Rong and Sang, Nong},
  booktitle={IEEE/CVF Conference on Computer Vision and Pattern Recognition},
  pages={19948--19957},
  year={2022}
}

@ARTICLE{9795869,
  author={Sun, Zehua and Ke, Qiuhong and Rahmani, Hossein and Bennamoun, Mohammed and Wang, Gang and Liu, Jun},
  journal={IEEE Transactions on Pattern Analysis and Machine Intelligence}, 
  title={Human Action Recognition From Various Data Modalities: A Review}, 
  year={2023},
  volume={45},
  number={3},
  pages={3200-3225},
  }

@article{snell2017prototypical,
  title={Prototypical networks for few-shot learning},
  author={Snell, Jake and Swersky, Kevin and Zemel, Richard},
  journal={Advances in Neural Information Processing Systems},
  volume={30},
  year={2017}
}

@article{huang2024matching,
  title={Matching compound prototypes for few-shot action recognition},
  author={Huang, Yifei and Yang, Lijin and Chen, Guo and Zhang, Hongjie and Lu, Feng and Sato, Yoichi},
  journal={International Journal of Computer Vision},
  volume={132},
  number={9},
  pages={3977--4002},
  year={2024},
}

@inproceedings{radford2021learning,
  title={Learning transferable visual models from natural language supervision},
  author={Radford, Alec and Kim, Jong Wook and Hallacy, Chris and Ramesh, Aditya and Goh, Gabriel and Agarwal, Sandhini and Sastry, Girish and Askell, Amanda and Mishkin, Pamela and Clark, Jack and others},
  booktitle={International Conference on Machine Learning},
  pages={8748--8763},
  year={2021},
  organization={PmLR}
}

@article{gao2021temporal,
  title={Temporal-attentive covariance pooling networks for video recognition},
  author={Gao, Zilin and Wang, Qilong and Zhang, Bingbing and Hu, Qinghua and Li, Peihua},
  journal={Advances in Neural Information Processing Systems},
  volume={34},
  pages={13587--13598},
  year={2021}
}

@article{wang2024clip,
  title={Clip-guided prototype modulating for few-shot action recognition},
  author={Wang, Xiang and Zhang, Shiwei and Cen, Jun and Gao, Changxin and Zhang, Yingya and Zhao, Deli and Sang, Nong},
  journal={International Journal of Computer Vision},
  volume={132},
  number={6},
  pages={1899--1912},
  year={2024},
  publisher={Springer}
}

@inproceedings{tran2015learning,
  title={Learning spatiotemporal features with 3d convolutional networks},
  author={Tran, Du and Bourdev, Lubomir and Fergus, Rob and Torresani, Lorenzo and Paluri, Manohar},
  booktitle={IEEE International Conference on Computer Vision},
  pages={4489--4497},
  year={2015}
}

@article{qu2024mvp,
  title={MVP-Shot: Multi-Velocity Progressive-Alignment Framework for Few-Shot Action Recognition},
  author={Qu, Hongyu and Yan, Rui and Shu, Xiangbo and Gao, Hailiang and Huang, Peng and Xie, Guo-Sen},
  journal={arXiv preprint arXiv:2405.02077},
  year={2024}
}

@article{li2024frame,
  title={Frame Order Matters: A Temporal Sequence-Aware Model for Few-Shot Action Recognition},
  author={Li, Bozheng and Liu, Mushui and Wang, Gaoang and Yu, Yunlong},
  journal={arXiv preprint arXiv:2408.12475},
  year={2024}
}

@inproceedings{huang2022compound,
  title={Compound prototype matching for few-shot action recognition},
  author={Huang, Yifei and Yang, Lijin and Sato, Yoichi},
  booktitle={European Conference on Computer Vision},
  pages={351--368},
  year={2022},
}

@INPROCEEDINGS{kuehne2011hmdb,
  author={Kuehne, H. and Jhuang, H. and Garrote, E. and Poggio, T. and Serre, T.},
  booktitle={2011 International Conference on Computer Vision}, 
  title={HMDB: A large video database for human motion recognition}, 
  year={2011},
  volume={},
  number={},
  pages={2556-2563},}

@article{soomro2012ucf101,
  title={UCF101: A dataset of 101 human actions classes from videos in the wild},
  author={Soomro, Khurram and Zamir, Amir Roshan and Shah, Mubarak},
  journal={arXiv preprint arXiv:1212.0402},
  year={2012}
}

@INPROCEEDINGS {carreira2017quo,
author = { Carreira, Joao and Zisserman, Andrew },
booktitle = {Computer Vision and Pattern Recognition},
title = {{ Quo Vadis, Action Recognition? A New Model and the Kinetics Dataset }},
year = {2017},
volume = {},
ISSN = {1063-6919},
pages = {4724-4733},
}

@inproceedings{vinyals2016matching,
  author={Vinyals, Oriol and Blundell, Charles and Lillicrap, Timothy and Kavukcuoglu, Koray and Wierstra, Daan},
  booktitle={Advances in Neural Information Processing Systems},
  title={Matching networks for one shot learning},
  pages = {3637--3645},
  volume={29},
  year={2016}
}

@inproceedings{perrett2021temporal,
  title={Temporal-relational crosstransformers for few-shot action recognition},
  author={Perrett, Toby and Masullo, Alessandro and Burghardt, Tilo and Mirmehdi, Majid and Damen, Dima},
  booktitle={Computer Vision and Pattern Recognition},
  pages={475--484},
  year={2021}
}

@inproceedings{dao2024transformers,
author = {Dao, Tri and Gu, Albert},
title = {Transformers are SSMs: generalized models and efficient algorithms through structured state space duality},
year = {2024},
booktitle = {International Conference on Machine Learning},
pages={10041--10071},
}

@inproceedings{fu2020depth,
author = {Fu, Yuqian and Zhang, Li and Wang, Junke and Fu, Yanwei and Jiang, Yu-Gang},
title = {Depth Guided Adaptive Meta-Fusion Network for Few-shot Video Recognition},
year = {2020},
booktitle = {ACM International Conference on Multimedia},
pages = {1142–1151},
numpages = {10},
}

@article{lee2025temporal,
  title={Temporal Alignment-Free Video Matching for Few-shot Action Recognition},
  author={Lee, SuBeen and Moon, WonJun and Seong, Hyun Seok and Heo, Jae-Pil},
  journal={arXiv preprint arXiv:2504.05956},
  year={2025}
}

@inproceedings{goyal2017something,
  title={The" something something" video database for learning and evaluating visual common sense},
  author={Goyal, Raghav and Ebrahimi Kahou, Samira and Michalski, Vincent and Materzynska, Joanna and Westphal, Susanne and Kim, Heuna and Haenel, Valentin and Fruend, Ingo and Yianilos, Peter and Mueller-Freitag, Moritz and others},
  booktitle={IEEE International Conference on Computer Vision},
  pages={5842--5850},
  year={2017}
}

@inproceedings{
gu2023mamba,
title={Mamba: Linear-Time Sequence Modeling with Selective State Spaces},
author={Albert Gu and Tri Dao},
booktitle={First Conference on Language Modeling},
year={2024}
}

@inproceedings{wang2021semantic,
  title={Semantic-guided relation propagation network for few-shot action recognition},
  author={Wang, Xiao and Ye, Weirong and Qi, Zhongang and Zhao, Xun and Wang, Guangge and Shan, Ying and Wang, Hanzi},
  booktitle={ACM International Conference on Multimedia},
  pages={816--825},
  year={2021}
}

@inproceedings{wang2023molo,
  title={Molo: Motion-augmented long-short contrastive learning for few-shot action recognition},
  author={Wang, Xiang and Zhang, Shiwei and Qing, Zhiwu and Gao, Changxin and Zhang, Yingya and Zhao, Deli and Sang, Nong},
  booktitle={IEEE/CVF Conference on Computer Vision and Pattern Recognition},
  pages={18011--18021},
  year={2023}
}

@inproceedings{yu2023multi,
  title={Multi-speed global contextual subspace matching for few-shot action recognition},
  author={Yu, Tianwei and Chen, Peng and Dang, Yuanjie and Huan, Ruohong and Liang, Ronghua},
  booktitle={ACM International Conference on Multimedia},
  pages={2344--2352},
  year={2023}

}

@inproceedings{zhang2023importance,
  title={On the importance of spatial relations for few-shot action recognition},
  author={Zhang, Yilun and Fu, Yuqian and Ma, Xingjun and Qi, Lizhe and Chen, Jingjing and Wu, Zuxuan and Jiang, Yu-Gang},
  booktitle={ACM International Conference on Multimedia},
  pages={2243--2251},
  year={2023}
}

@article{gu2021combining,
  title={Combining recurrent, convolutional, and continuous-time models with linear state space layers},
  author={Gu, Albert and Johnson, Isys and Goel, Karan and Saab, Khaled and Dao, Tri and Rudra, Atri and R{\'e}, Christopher},
  journal={Advances in Neural Information Processing Systems},
  volume={34},
  pages={572--585},
  year={2021}
}

@inproceedings{gu2021efficiently,
title={Efficiently Modeling Long Sequences with Structured State Spaces},
author={Albert Gu and Karan Goel and Christopher Re},
booktitle={International Conference on Learning Representations},
year={2022},
}

@inproceedings{vaswani2017attention,
author = {Vaswani, Ashish and Shazeer, Noam and Parmar, Niki and Uszkoreit, Jakob and Jones, Llion and Gomez, Aidan N. and Kaiser, \L{}ukasz and Polosukhin, Illia},
title = {Attention is all you need},
year = {2017},
booktitle = {Proceedings of the 31st International Conference on Neural Information Processing Systems},
pages = {6000--6010},
numpages = {11},
}

@article{yu2024mambaout,
  title={Mambaout: Do we really need mamba for vision?},
  author={Yu, Weihao and Wang, Xinchao},
  journal={arXiv preprint arXiv:2405.07992},
  year={2024}
}

@inproceedings{bishay2019tarn,
  author={Mina Bishay and Georgios Zoumpourlis and Ioannis Patras},
  title={TARN: Temporal Attentive Relation Network for Few-Shot and Zero-Shot Action Recognition},
  year={2019},
  pages={154},
  booktitle={British Machine Vision Conference},
}

@inproceedings{zhu2024visionmambaefficientvisual,
author = {Zhu, Lianghui and Liao, Bencheng and Zhang, Qian and Wang, Xinlong and Liu, Wenyu and Wang, Xinggang},
title = {Vision mamba: efficient visual representation learning with bidirectional state space model},
year = {2024},
booktitle = {International Conference on Machine Learning},
pages = {62429--62442},
}

@article{liu2024vmamba,
  title={Vmamba: Visual state space model},
  author={Liu, Yue and Tian, Yunjie and Zhao, Yuzhong and Yu, Hongtian and Xie, Lingxi and Wang, Yaowei and Ye, Qixiang and Jiao, Jianbin and Liu, Yunfan},
  journal={Advances in Neural Information Processing Systems},
  volume={37},
  pages={103031--103063},
  year={2024}
}

@inproceedings{li2024videomamba,
  title={Videomamba: State space model for efficient video understanding},
  author={Li, Kunchang and Li, Xinhao and Wang, Yi and He, Yinan and Wang, Yali and Wang, Limin and Qiao, Yu},
  booktitle={European Conference on Computer Vision},
  pages={237--255},
  year={2024},
}

@article{liu2022spatial,
  title={Spatial channel attention for deep convolutional neural networks},
  author={Liu, Tonglai and Luo, Ronghai and Xu, Longqin and Feng, Dachun and Cao, Liang and Liu, Shuangyin and Guo, Jianjun},
  journal={Mathematics},
  volume={10},
  number={10},
  pages={1750},
  year={2022},
}

@inproceedings{nie2024slowfocus,
  title={SlowFocus: Enhancing Fine-grained Temporal Understanding in Video LLM},
  author={Nie, Ming and Ding, Dan and Wang, Chunwei and Guo, Yuanfan and Han, Jianhua and Xu, Hang and Zhang, Li},
  booktitle={Advances in Neural Information Processing Systems},
  year={2024}
}

@inproceedings{zhang2020few,
  title={Few-shot action recognition with permutation-invariant attention},
  author={Zhang, Hongguang and Zhang, Li and Qi, Xiaojuan and Li, Hongdong and Torr, Philip HS and Koniusz, Piotr},
  booktitle={European Conference on Computer Vision},
  pages={525--542},
  year={2020},
}

@inproceedings{he2016deep,
  title={Deep residual learning for image recognition},
  author={He, Kaiming and Zhang, Xiangyu and Ren, Shaoqing and Sun, Jian},
  booktitle={Computer Vision and Pattern Recognition},
  pages={770--778},
  year={2016}
}

@inproceedings{dosovitskiy2020image,
title={An Image is Worth 16x16 Words: Transformers for Image Recognition at Scale},
author={Alexey Dosovitskiy and Lucas Beyer and Alexander Kolesnikov and Dirk Weissenborn and Xiaohua Zhai and Thomas Unterthiner and Mostafa Dehghani and Matthias Minderer and Georg Heigold and Sylvain Gelly and Jakob Uszkoreit and Neil Houlsby},
booktitle={International Conference on Learning Representations},
year={2021}
}

@article{achiam2023gpt,
  title={Gpt-4 technical report},
  author={Achiam, Josh and Adler, Steven and Agarwal, Sandhini and Ahmad, Lama and Akkaya, Ilge and Aleman, Florencia Leoni and Almeida, Diogo and Altenschmidt, Janko and Altman, Sam and Anadkat, Shyamal and others},
  journal={arXiv preprint arXiv:2303.08774},
  year={2023}
}

@misc{openai2024gpt4o,
  author       = {{OpenAI}},
  title        = {Hello GPT-4o},
  year         = {2024},
  howpublished = {\url{https://openai.com/index/hello-gpt-4o}},
  note         = {Accessed: 2025-08-05}
}

@misc{deepmind2025gemini,
  author       = {{Google DeepMind}},
  title        = {Gemini 2.5: Our most intelligent AI model},
  year         = {2025},
  howpublished = {\url{https://blog.google/technology/google-deepmind/gemini-model-thinking-updates-march-2025/}},
  note         = {Accessed: 2025-08-05}
}

@article{su2024reallocating,
  title={Reallocating and Evolving General Knowledge for Few-Shot Learning},
  author={Su, Yuling and Liu, Xueliang and Huang, Zhen and He, Jun and Hong, Richang and Wang, Meng},
  journal={IEEE Transactions on Circuits and Systems for Video Technology},
  year={2024},
  publisher={IEEE}
}

@article{yin2024hierarchy,
  title={Hierarchy-aware interactive prompt learning for few-shot classification},
  author={Yin, Xiaotian and Wu, Jiamin and Yang, Wenfei and Zhou, Xu and Zhang, Shifeng and Zhang, Tianzhu},
  journal={IEEE Transactions on Circuits and Systems for Video Technology},
  year={2024},
  publisher={IEEE}
}

@inproceedings{farina2025rethinking,
  title={Rethinking Few-Shot Adaptation of Vision-Language Models in Two Stages},
  author={Farina, Matteo and Mancini, Massimiliano and Iacca, Giovanni and Ricci, Elisa},
  booktitle={Proceedings of the Computer Vision and Pattern Recognition Conference},
  pages={29989--29998},
  year={2025}
}

@article{zhang2024few,
  title={Few-shot cross-domain object detection with instance-level prototype-based meta-learning},
  author={Zhang, Lin and Zhang, Bo and Shi, Botian and Fan, Jiayuan and Chen, Tao},
  journal={IEEE Transactions on Circuits and Systems for Video Technology},
  volume={34},
  number={10},
  pages={9078--9089},
  year={2024},
  publisher={IEEE}
}

@inproceedings{jung2022few,
  title={Few-shot metric learning: Online adaptation of embedding for retrieval},
  author={Jung, Deunsol and Kang, Dahyun and Kwak, Suha and Cho, Minsu},
  booktitle={Proceedings of the Asian Conference on Computer Vision},
  pages={1875--1891},
  year={2022}
}

@article{lin2024revisiting,
  title={Revisiting few-shot learning from a causal perspective},
  author={Lin, Guoliang and Xu, Yongheng and Lai, Hanjiang and Yin, Jian},
  journal={IEEE Transactions on Knowledge and Data Engineering},
  volume={36},
  number={11},
  pages={6908--6919},
  year={2024},
  publisher={IEEE}
}

@article{zhou2024meta,
  title={Meta-exploiting frequency prior for cross-domain few-shot learning},
  author={Zhou, Fei and Wang, Peng and Zhang, Lei and Chen, Zhenghua and Wei, Wei and Ding, Chen and Lin, Guosheng and Zhang, Yanning},
  journal={Advances in Neural Information Processing Systems},
  volume={37},
  pages={116783--116814},
  year={2024}
}

@article{wang2024stability,
  title={On the stability and generalization of meta-learning},
  author={Wang, Yunjuan and Arora, Raman},
  journal={Advances in Neural Information Processing Systems},
  volume={37},
  pages={83665--83710},
  year={2024}
}

@inproceedings{cao2024task,
  title={Task-adapter: Task-specific adaptation of image models for few-shot action recognition},
  author={Cao, Congqi and Zhang, Yueran and Yu, Yating and Lv, Qinyi and Min, Lingtong and Zhang, Yanning},
  booktitle={Proceedings of the 32nd ACM International Conference on Multimedia},
  pages={9038--9047},
  year={2024}
}

@article{lu2024cross,
  title={Cross-modal contrastive pre-training for few-shot skeleton action recognition},
  author={Lu, Mingqi and Yang, Siyuan and Lu, Xiaobo and Liu, Jun},
  journal={IEEE Transactions on Circuits and Systems for Video Technology},
  volume={34},
  number={10},
  pages={9798--9807},
  year={2024},
  publisher={IEEE}
}

@article{wang2024few,
  title={Few-shot action recognition via multi-view representation learning},
  author={Wang, Xiao and Lu, Yang and Yu, Wanchuan and Pang, Yanwei and Wang, Hanzi},
  journal={IEEE Transactions on Circuits and Systems for Video Technology},
  volume={34},
  number={9},
  pages={8522--8535},
  year={2024},
  publisher={IEEE}
}

@inproceedings{wang2023selective,
  title={Selective structured state-spaces for long-form video understanding},
  author={Wang, Jue and Zhu, Wentao and Wang, Pichao and Yu, Xiang and Liu, Linda and Omar, Mohamed and Hamid, Raffay},
  booktitle={Proceedings of the IEEE/CVF Conference on Computer Vision and Pattern Recognition},
  pages={6387--6397},
  year={2023}
}

@inproceedings{xie2025mamba,
  title={Mamba-Adaptor: State Space Model Adaptor for Visual Recognition},
  author={Xie, Fei and Nie, Jiahao and Tang, Yujin and Zhang, Wenkang and Zhao, Hongshen},
  booktitle={Proceedings of the Computer Vision and Pattern Recognition Conference},
  pages={20124--20134},
  year={2025}
}

@inproceedings{hatamizadeh2025mambavision,
  title={Mambavision: A hybrid mamba-transformer vision backbone},
  author={Hatamizadeh, Ali and Kautz, Jan},
  booktitle={Proceedings of the Computer Vision and Pattern Recognition Conference},
  pages={25261--25270},
  year={2025}
}

@inproceedings{shaker2025groupmamba,
  title={GroupMamba: Efficient Group-Based Visual State Space Model},
  author={Shaker, Abdelrahman and Wasim, Syed Talal and Khan, Salman and Gall, Juergen and Khan, Fahad Shahbaz},
  booktitle={Proceedings of the Computer Vision and Pattern Recognition Conference},
  pages={14912--14922},
  year={2025}
}

@article{kingma2014adam,
  title={Adam: A method for stochastic optimization},
  author={Kingma, Diederik P and Ba, Jimmy},
  journal={arXiv preprint arXiv:1412.6980},
  year={2014}
}

@article{liu2025motion,
  title={Motion-consistent representation learning for uav-based action recognition},
  author={Liu, Wenxuan and Zhong, Xian and Dai, Yihan and Jia, Xuemei and Wang, Zheng and Satoh, Shin’Ichi},
  journal={IEEE Transactions on Intelligent Transportation Systems},
  year={2025},
  publisher={IEEE}
}

@article{liu2023dual,
  title={Dual-recommendation disentanglement network for view fuzz in action recognition},
  author={Liu, Wenxuan and Zhong, Xian and Zhou, Zhuo and Jiang, Kui and Wang, Zheng and Lin, Chia-Wen},
  journal={IEEE Transactions on Image Processing},
  volume={32},
  pages={2719--2733},
  year={2023},
  publisher={IEEE}
}

@misc{liu2026unify,
      title={Unify the Views: View-Consistent Prototype Learning for Few-Shot Segmentation}, 
      author={Hongli Liu and Yu Wang and Shengjie Zhao},
      year={2026},
      eprint={2603.05952},
      archivePrefix={arXiv},
      primaryClass={cs.CV},
      url={https://arxiv.org/abs/2603.05952}, 
}

@article{tang2022learning,
  title={Learning attention-guided pyramidal features for few-shot fine-grained recognition},
  author={Tang, Hao and Yuan, Chengcheng and Li, Zechao and Tang, Jinhui},
  journal={Pattern Recognition},
  volume={130},
  pages={108792},
  year={2022},
  publisher={Elsevier}
}

@inproceedings{tang2020blockmix,
  title={Blockmix: meta regularization and self-calibrated inference for metric-based meta-learning},
  author={Tang, Hao and Li, Zechao and Peng, Zhimao and Tang, Jinhui},
  booktitle={Proceedings of the 28th ACM international conference on multimedia},
  pages={610--618},
  year={2020}
}

@inproceedings{tang2023m3net,
  title={M3net: multi-view encoding, matching, and fusion for few-shot fine-grained action recognition},
  author={Tang, Hao and Liu, Jun and Yan, Shuanglin and Yan, Rui and Li, Zechao and Tang, Jinhui},
  booktitle={Proceedings of the 31st ACM international conference on multimedia},
  pages={1719--1728},
  year={2023}
}

@inproceedings{tang2025connecting,
author = {Tang, Hao and He, Shengfeng and Qin, Jing},
title = {Connecting giants: synergistic knowledge transfer of large multimodal models for few-shot learning},
year = {2025},
booktitle = {Proceedings of the Thirty-Fourth International Joint Conference on Artificial Intelligence},
pages={6227--6235},
series = {IJCAI '25}
}

@article{tang2026cross,
    title={Cross-modal Proxy Evolving for OOD Detection with Vision-Language Models}, 
    number={18}, 
    journal={Proceedings of the AAAI Conference on Artificial Intelligence}, 
    author={Tang, Hao and Liu, Yu and Yan, Shuanglin and Shen, Fei and He, Shengfeng and Qin, Jing}, 
    year={2026}, 
    pages={15770-15778} 
}

\begin{IEEEbiography}[{\includegraphics[width=1in,height=1.25in,clip,keepaspectratio]{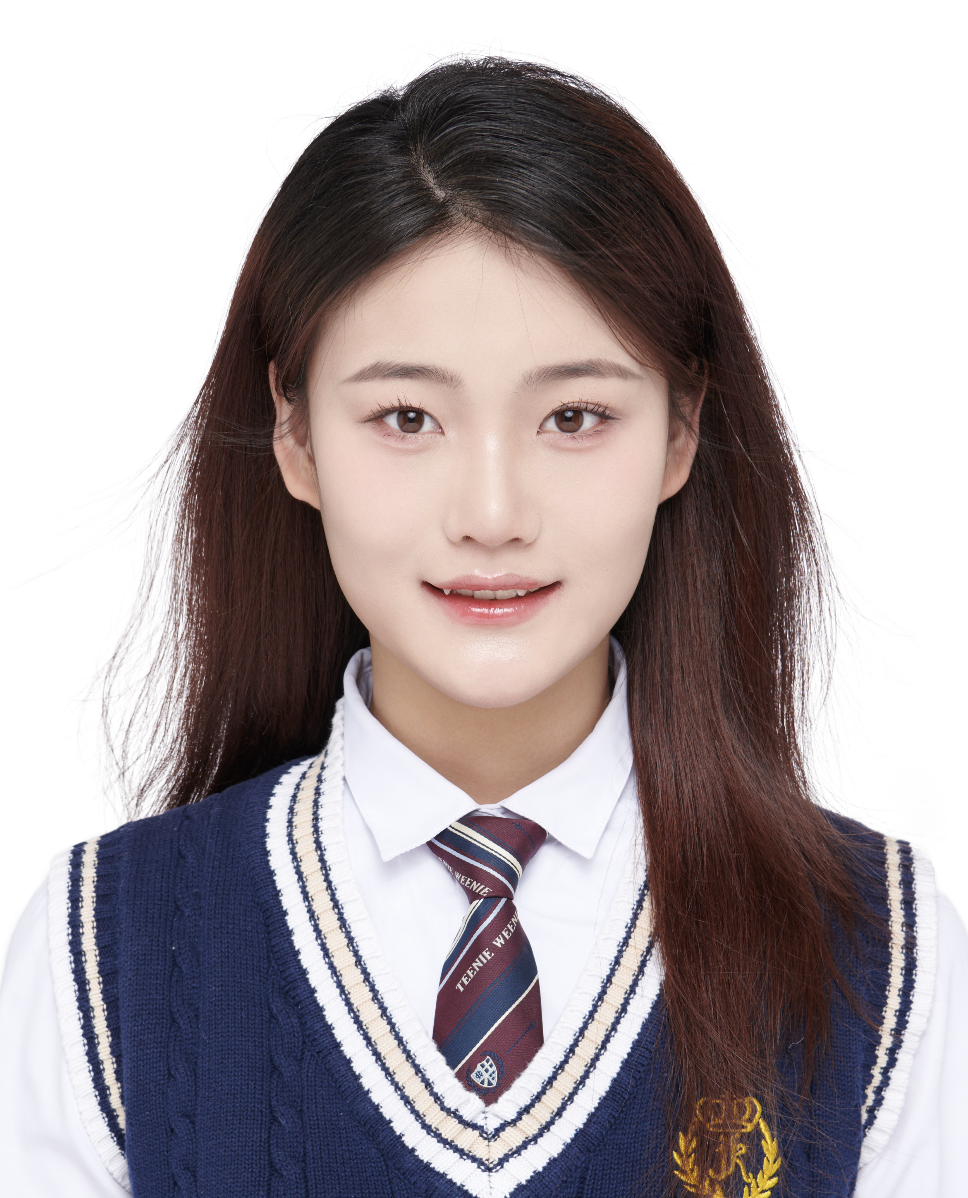}}]{Hongli Liu}
Hongli Liu is currently pursuing the Ph.D. degree with the School of Computer Science, Tongji University, Shanghai, China. Her main research interests include computer vision, few-shot learning, and multimodal video understanding.
\end{IEEEbiography}
\begin{IEEEbiography}[{\includegraphics[width=1in,height=1.25in,clip]{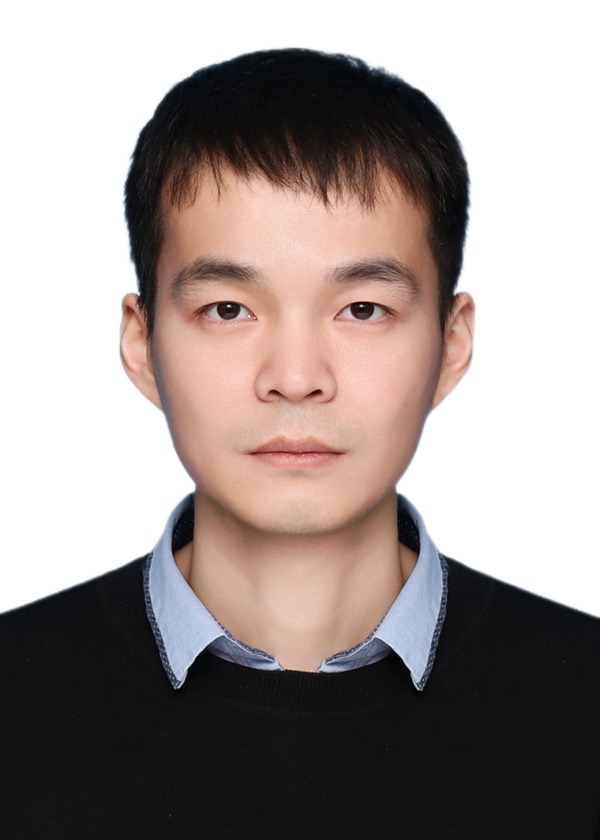}}]{Yu Wang} [Member IEEE] received the PhD degree from Tongji University, Shanghai, China, in 2022. Since 2023, he has been an assistant professor at Tongji University, Shanghai, China. His current research interests include artificial intelligence, computer vision, multi-modal learning, and video understanding and generation. As the first author or corresponding author, he has published more than 20 papers at top conferences and journals. Currently, he serves as the reviewer of CVPR, ICCV, ICLR, ACM MM, TNNLS, TMM, TCSVT, T-ITS, etc. He was also an outstanding reviewer of CVPR 2025. He has won multiple challenge championship honors, including CVPR 2025 and ICDAR 2023.
\end{IEEEbiography}
\begin{IEEEbiography}[{\includegraphics[width=1in,height=1.25in,clip]{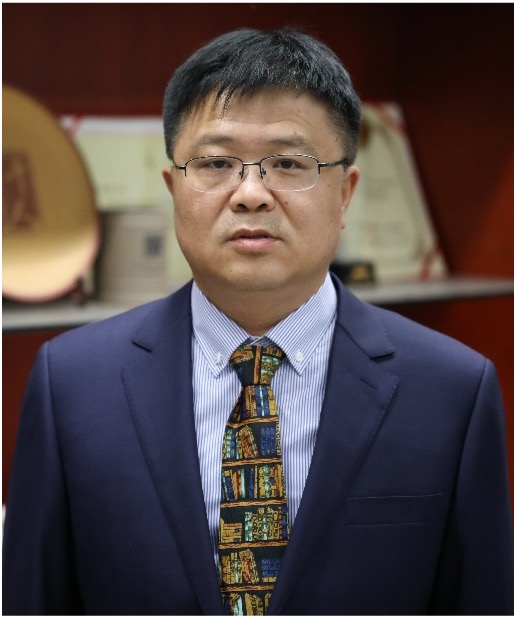}}]{Shengjie Zhao} [Senior Member, IEEE] received his B.S. degree in electrical engineering from the University of Science and Technology of China, Hefei, in 1988; his M.S. degree in electrical and computer engineering from China Aerospace Institute, Beijing, in 1991; and his Ph.D. degree in electrical and computer engineering from Texas A\&M University, College Station, Texas, in 2004. His research interests include artificial intelligence, computer vision, multi-modal learning, and image processing. He is an academician of the International Eurasian Academy of Sciences (IEAS) and a Fellow of the Thousand Talents program of China.
\end{IEEEbiography}

\vfill

\end{document}